\documentclass[]{lbw}
\definecolor{metablue}{HTML}{1772B4}
\definecolor{tablerowcolor}{RGB}{235,245,252}
\definecolor{tablerowcolor1}{RGB}{235,248,235}
\definecolor{mygreen}{RGB}{84,130,53}
\setabstractbrandtext{}
\setabstractbrandcolor{metablue}
\setabstractbrandlogo{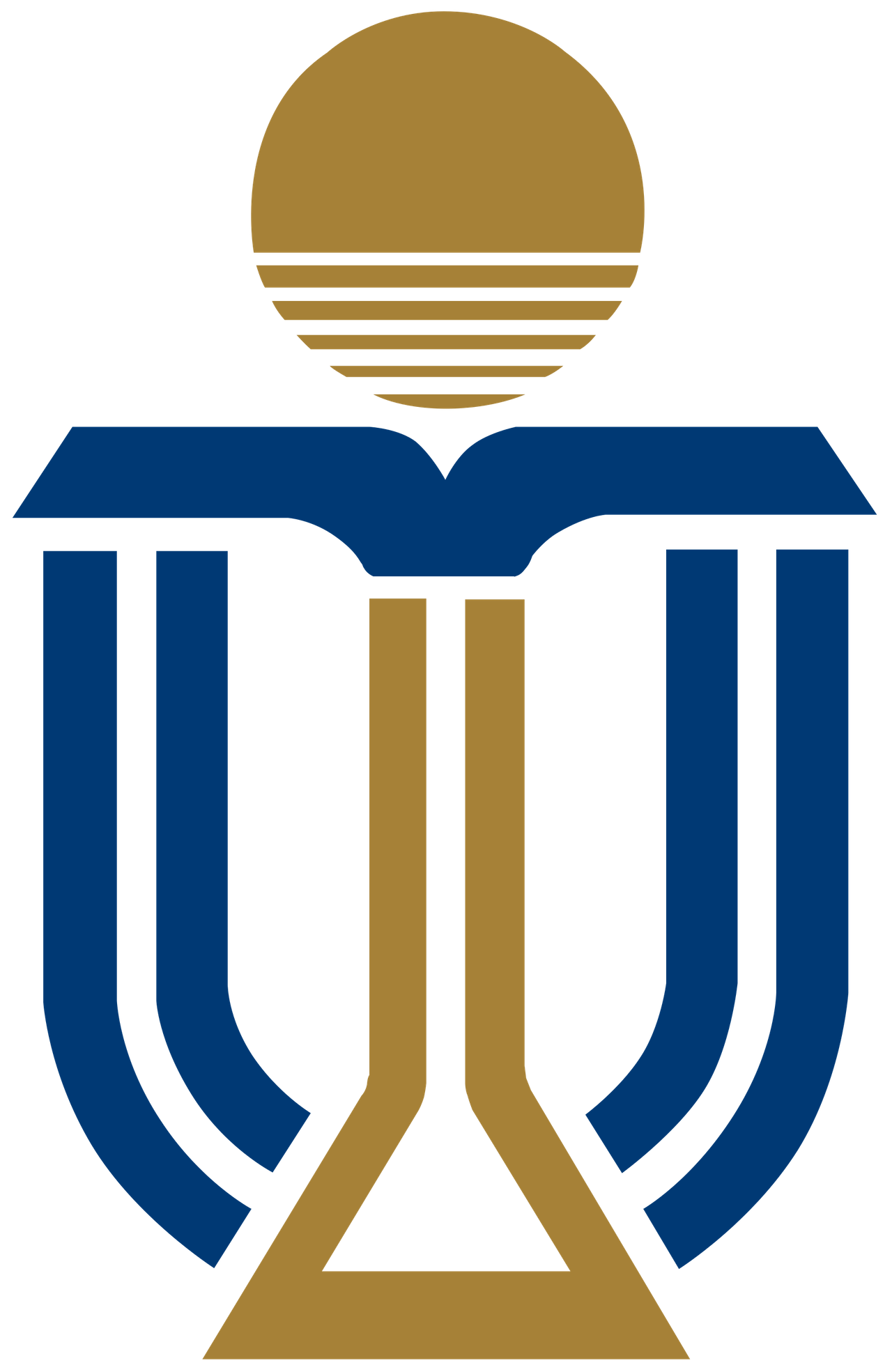}
\setabstractbrandlogoheight{1cm}

\usepackage{amsmath,amssymb,amsfonts}
\usepackage{mathtools}
\usepackage{amsthm}
\usepackage{bbm}
\usepackage{bm}
\usepackage{subfigure}
\usepackage{float}
\usepackage{adjustbox}
\usepackage{colortbl}
\usepackage{arydshln}
\usepackage{algorithm,algpseudocode}
\usepackage{algorithmicx}
\usepackage{wrapfig}
\usepackage{pifont}
\usepackage{bbding}
\usepackage{xspace}
\usepackage{multicol}
\usepackage{array}
\usepackage{multirow}
\usepackage{makecell}
\usepackage{pgf}
\usepackage{enumitem}
\usepackage{tabularx}
\usepackage{longtable}
\usepackage{xurl}
\usepackage{nicefrac}

\newcolumntype{Y}{>{\raggedright\arraybackslash}X}
\newcolumntype{P}[1]{>{\raggedright\arraybackslash}p{#1}}

\setlist[itemize]{leftmargin=1.2em,itemsep=1pt,topsep=2pt}
\setlist[enumerate]{leftmargin=1.4em,itemsep=1pt,topsep=2pt}
\definecolor{myblue}{HTML}{76C7E5}
\newcommand{\heatcell}[3]{%
    \pgfmathsetmacro{\percent}{(#1 - #2) / (#3 - #2) * 50 + 10}%
    \edef\temp{\noexpand\cellcolor{myblue!\percent}\relax \noexpand\makebox[2.2em][r]{#1}}%
    \temp
}
\newcommand{\heatcellb}[3]{%
    \pgfmathsetmacro{\percent}{(#1 - #2) / (#3 - #2) * 50 + 10}%
    \edef\temp{\noexpand\cellcolor{myblue!\percent}\relax \noexpand\makebox[2.2em][r]{\noexpand\textbf{#1}}}%
    \temp
}
\newcommand{\cmark}{\textcolor{green!60!black}{\ding{51}}}
\newcommand{\xmark}{\textcolor{red}{\ding{55}}}
\newcommand{\ours}{\textsc{GenEvA}\xspace}

\makeatletter
\DeclareRobustCommand\onedot{\futurelet\@let@token\@onedot}
\def\@onedot{\ifx\@let@token.\else.\null\fi\xspace}

\makeatother

\title{Beyond Frame Selection: Generative Latent Evidence Aggregation for Long-Video Understanding}

\author[1\dagger]{Bowen Liu}
\author[3\dagger]{Shuning Wang}
\author[4]{Xinpeng Ding}
\author[2]{Zhiheng Wu}
\author[1]{Bodong Du}
\author[1]{Xiaomeng Li}
\affiliation[1]{The Hong Kong University of Science and Technology}
\affiliation[2]{Baidu Inc.}
\affiliation[3]{Alibaba Group}
\affiliation[4]{Xidian University}
\contribution[\dagger]{Equal Contribution}
\correspondence{Xiaomeng Li}

\abstract{%
Long-video understanding commonly compresses videos into a small set of frames
or visual tokens for answer generation. Existing compact pipelines focus on
retaining relevant visual content as explicit evidence. Yet making evidence
available does not ensure that complementary cues across moments are integrated
for answering. Our key idea is to organize selected frames into query-relevant
cross-frame evidence before generation. We formulate this post-selection stage
as a \emph{latent evidence interface} and instantiate it with \ours
(\textbf{Gen}erative Latent \textbf{Ev}idence \textbf{A}ggregation), a
distribution-guided latent evidence aggregation framework. Specifically,
\ours uses a query-conditioned evidence distribution to focus aggregation on
relevant frames, forming compact cross-frame latent evidence from their
frame-specific information. Since cross-frame integration is not always needed,
the same distribution determines whether to insert this latent complement.
Across four benchmarks and two Video-MLLM backbones, \ours consistently
improves matched-frame baselines. At 8 frames, it raises the four-benchmark
LLaVA-Video average by $+5.2$ points and Qwen2.5-VL accuracy on LVBench by
$+10.1$ points. These gains require only $0.11\%$--$0.40\%$ average video-token
overhead; analyses further show task-aware allocation and benefits from
Adaptive Evidence Invocation.
}

\begin{document}

\maketitle

\section{Introduction}

Long-video understanding requires Video-MLLMs to answer questions when relevant
evidence is sparse, temporally distant, and unevenly distributed across a
video. Under practical visual budgets, inference must reduce the video to a
compact context. Success in this regime therefore depends on two distinct
stages: retaining potentially relevant frames and converting their visual
content into evidence that the generator can use.

Most existing methods focus on the input side of this pipeline by constructing
\emph{explicit evidence} before generation. Long-context models extend temporal
coverage~\cite{zhang2024long,chen2025longvila}; efficient architectures and
token-compression methods reduce visual cost~\cite{ren2025vamba,shen2024longvu,
shu2025video,liu2025video}; and adaptive samplers retrieve query-related
frames~\cite{li2026divide,zou2026videobrain}. Collectively, these approaches
determine what visual content reaches the generator and in what explicit form.
They leave a consequential assumption largely unexamined: once relevant visual
evidence is retained, the Video-MLLM will use it effectively.

\begin{wrapfigure}{r}{0.50\textwidth}
  \centering
  \includegraphics[width=\linewidth]{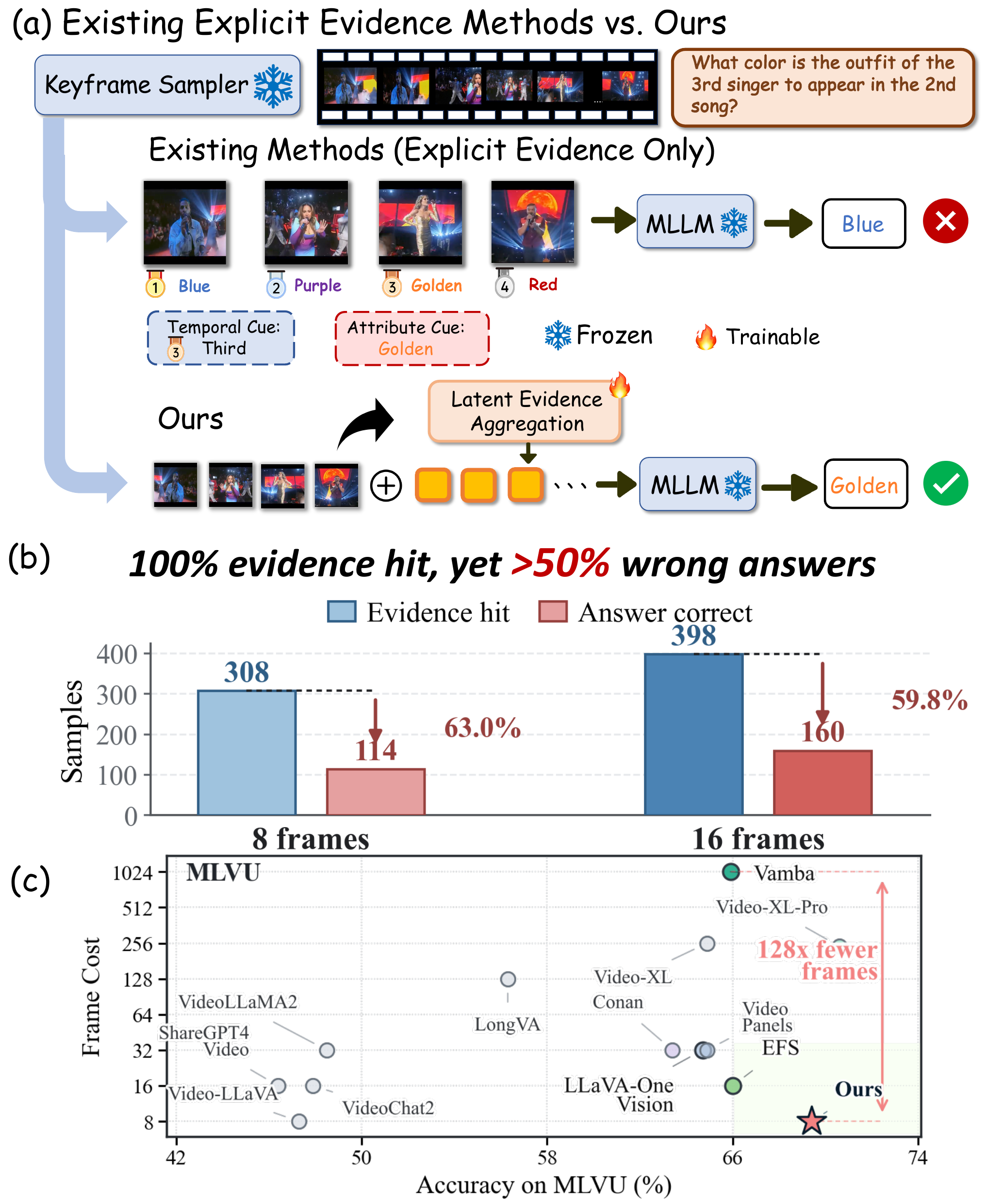}
  \caption{\textbf{Motivation and cost-effectiveness of \ours{}.}
  (a--b) Qualitative and quantitative evidence exposes a post-selection
  evidence-use bottleneck. (c) The reported MLVU landscape illustrates the
  cost-effectiveness of \ours{} under a compact frame budget.}
  \label{fig:intro_motivation}
\end{wrapfigure}

Evidence availability, however, does not guarantee evidence use. Recent studies
show that MLLMs may underuse available visual evidence and instead follow
language priors or spurious contextual cues~\cite{favero2024multimodal,
asadi2026mirage,benlevi2026mirageprobes,wu2025postalign,zhu2026visualflip,
morini2026looktwice,xiao2026vigil}. Figure~\ref{fig:intro_motivation}(a)
illustrates this failure: the retained frames jointly provide the temporal-order
and attribute cues needed for the answer, yet the explicit-only route predicts
blue instead of golden. Our LVBench diagnostic in
Figure~\ref{fig:intro_motivation}(b) shows the same pattern at scale. Even when
frames selected by the frozen temporal selector overlap annotated evidence, an
explicit-only Qwen2.5-VL remains incorrect on more than half of the questions
at both frame budgets. These findings expose a post-selection bottleneck:
evidence can reach the model without being reliably converted into an
answer-relevant representation. The challenge remains: \emph{How can a Video-MLLM construct
query-relevant latent evidence from selected frames to complement the explicit
visual context?}\\
We address this bottleneck through a \emph{latent evidence interface} between
frame selection and answer generation. The interface preserves selected frames
as explicit visual context while adding a compact latent channel that organizes
their query-relevant cues for generation. We instantiate it with \ours{}
(\textbf{Gen}erative Latent \textbf{Ev}idence \textbf{A}ggregation), a
distribution-guided latent evidence aggregation framework for compact
long-video understanding. Given frames returned by a frozen pretrained temporal
selector, \ours{} allocates a fixed latent-token budget across them and
aggregates frame-specific cues from frozen visual features into
query-conditioned latent evidence. Adaptive Evidence Invocation then uses the
dispersion of the resulting allocation as an intrinsic signal to select the
generation route for each query. The same allocation therefore governs both
latent capacity and evidence use without a separately trained router.

Across Video-MME, MLVU, LongVideoBench, and LVBench, \ours{} consistently
improves matched-frame baselines on two Video-MLLM backbones. At 8 frames, it
raises the LLaVA-Video four-benchmark average by $+5.2$ points and improves
Qwen2.5-VL by up to $+10.1$ points on LVBench. These gains use at most 32 latent
tokens, corresponding to only $0.11\%$--$0.40\%$ average video-token overhead.
Figure~\ref{fig:intro_motivation}(c) provides the broader reported MLVU
landscape: \ours{} reaches $69.4$ accuracy with 8 frames, while Vamba reports
$65.9$ with 1024 frames. Further analyses connect allocation dispersion to task
evidence structure and show that Adaptive Evidence Invocation outperforms Full
and Random invocation on average.

We summarize the contributions as follows:
\begin{itemize}
    \item We identify a post-selection evidence-use bottleneck and formulate
    the \textbf{latent evidence interface} as a missing stage between frame
    selection and answer generation.
    \item We propose \ours{}, a lightweight, distribution-guided latent
    evidence aggregation framework that realizes this interface through
    Evidence-Budget Allocation, Query-Aware Latent Evidence Aggregation, and
    Adaptive Evidence Invocation.
    \item We evaluate \ours{} on four long-video benchmarks and two Video-MLLM
    backbones, demonstrating consistent matched-frame gains, low token
    overhead, task-aware allocation patterns, and gains from Adaptive Evidence
    Invocation.
\end{itemize}

\section{Related Work}

\noindent\textbf{Long-video understanding.} Existing Video-MLLMs mainly make long visual inputs tractable before generation. Long-context models increase temporal coverage by scaling the context window or transferring long-context language modeling to video~\cite{zhang2024long,chen2025longvila}. Compact-context methods reduce visual cost with frame-level summaries, token compression, or hybrid sequence modeling~\cite{li2024llama,shen2024longvu,shu2025video,liu2025video,ren2025vamba}. Adaptive selection and input-organization methods choose query-relevant frames, optimize temporal sampling policies, impose causal or event-aware coverage, or repack long videos into panel-style visual inputs under a limited budget~\cite{li2026divide,tang2026tspo,zou2026videobrain,zhou2026reason,chen2026efs,doorenbos2026videopanels}. Reasoning-oriented Video-MLLMs further add tool use, multi-turn segment selection, reinforcement learning, multi-scale evidence reasoning, or visual-textual chain-of-thought~\cite{yang2026longvt,xie2026videomtr,feng2025videor1,ouyang2025conan,zhang2026vticot}. These methods improve either which visual context reaches the generator or how the generator reasons over the supplied context. \ours{} instead makes explicit a latent evidence interface after frame selection: it treats selected frames as candidate visual evidence and aggregates query-conditioned latent evidence from them.

\noindent\textbf{Evidence grounding and evidence use.} A parallel line of work asks whether MLLMs ground their answers in the intended visual evidence. Grounded MLLMs make this evidence explicit through regions, masks, video moments, or source citations~\cite{peng2023kosmos2,chen2023shikra,you2023ferret,rasheed2023glamm,li2024groundinggpt,lai2023lisa,yan2024visa,munasinghe2024videoglamm}. Evidence-centric benchmarks and inference methods further show that answer accuracy can hide weak visual dependence~\cite{zhu2026visualflip,azadani2026vistaqa,morini2026looktwice,asadi2026mirage,benlevi2026mirageprobes}. Recent MLLM studies also make the evidence interface more explicit by preserving answer-critical contexts, grounding temporal evidence, or aligning latent reasoning states with spatial-semantic evidence~\cite{liu2026moment,ahn2026evident,li2026keeping,cui2026ris}. Together, these studies highlight a gap between making visual evidence available and using it effectively. \ours{} targets this gap after frame selection by aggregating query-relevant cues across selected frames into a compact latent complement to the explicit visual context.

\section{Method}

\begin{figure}[t]
    \centering
    \includegraphics[width=\textwidth]{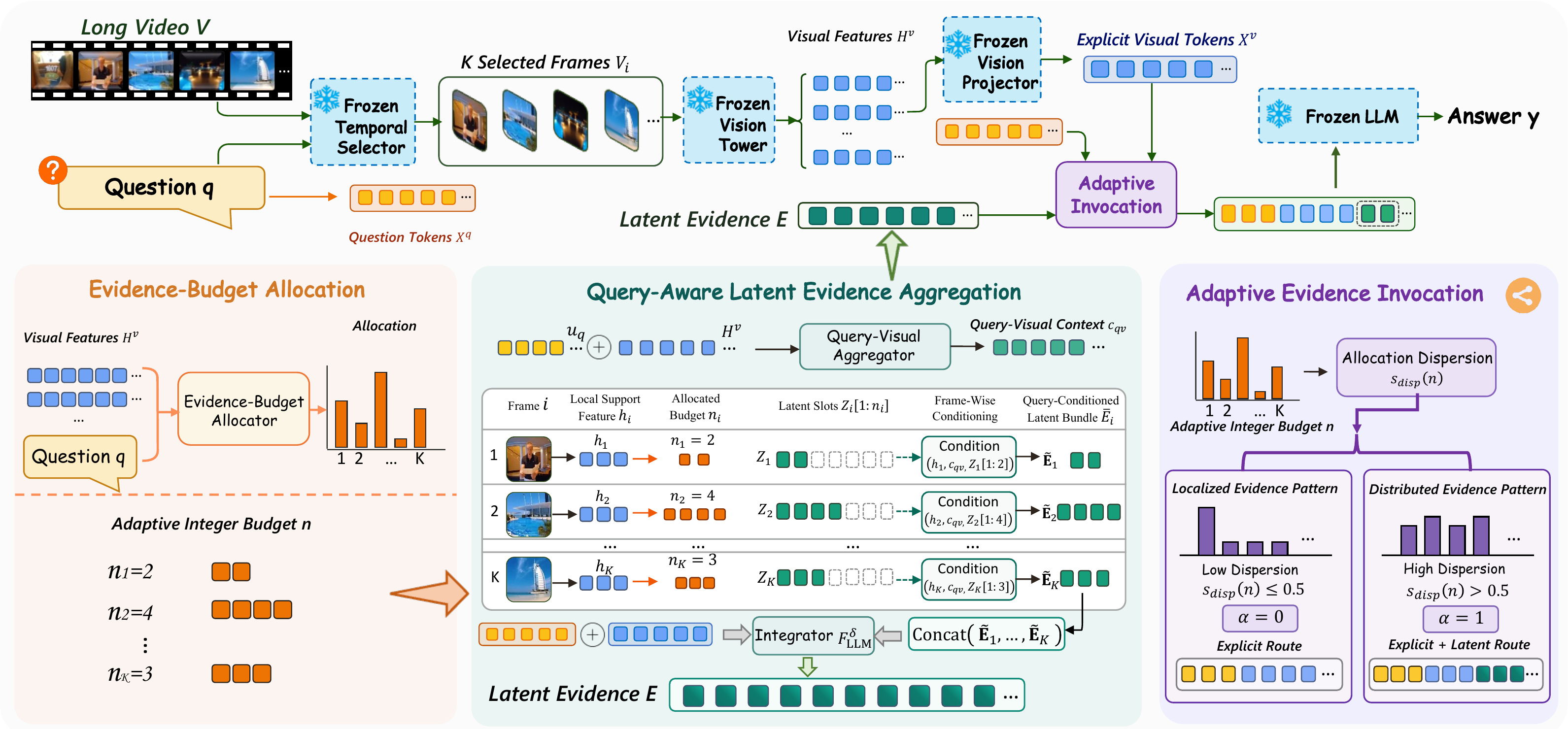}
    \caption{Overview of \ours{}. Given frames from a frozen temporal selector,
    \ours{} allocates frame-wise latent capacity, aggregates query-aware latent
    evidence, and applies Adaptive Evidence Invocation according to allocation
    dispersion.}
    \label{fig:method_overview}
\end{figure}

\subsection{Problem Formulation: Latent Evidence Interface}
\label{sec:method_overview}

Given a long video $V=\{v_t\}_{t=1}^{T}$ and a question $q$, compact
long-video QA generates an answer $y$ under a frame budget $K\ll T$. A frozen
pretrained temporal selector $\mathcal{L}_{\psi^{\star}}$ returns an index
sequence $\mathcal{I}=(t_1,\ldots,t_K)$, with $t_1<\cdots<t_K$, and the
corresponding selected-frame sequence
$V_{\mathcal{I}}=(v_{t_1},\ldots,v_{t_K})$. We index the selected frames in
temporal order, so position $i$ refers to $v_{t_i}$.

The frozen Video-MLLM contains a vision tower $F_{\mathrm{vis}}$, a visual
projector $P_{\mathrm{vis}}$, and a language model $F_{\mathrm{LLM}}$. Let
$\mathbf{H}^{v}=(\mathbf{H}^{v}_{1},\ldots,\mathbf{H}^{v}_{K})$ denote the
vision-tower tokens of the selected frames. We further denote the pre-LLM
question-token embeddings by $\mathbf{X}^{q}$ and their pooled representation by
$\mathbf{u}_{q}=\mathrm{Pool}(\mathbf{X}^{q})\in\mathbb{R}^{d_q}$.

\ours{} instantiates a post-selection latent evidence interface between
selection and generation. It preserves selected frames as explicit visual
context while constructing a compact latent complement for the generator.
Given a latent-token budget $B$ and a per-frame latent-basis capacity
$N_{\max}$, the interface answers three questions: how
much latent capacity each frame receives, what evidence fills that capacity,
and whether the resulting evidence is inserted for generation. The trainable
allocator and aggregation head form a lightweight evidence adapter, while
Adaptive Evidence Invocation selects the generation route with a deterministic
allocation-dispersion rule. Both inference routes
retain the explicit visual input; they differ in whether the aggregation module
is invoked and its latent evidence is inserted for answer generation. We assume
$2\le K<B$ and
$N_{\max}\ge B-K+1$, ensuring a nonempty residual budget and feasible
frame-wise allocations.
Figure~\ref{fig:method_overview} summarizes the resulting inference interface.

\subsection{Evidence-Budget Allocation}
\label{sec:evidence_budget_allocator}

The selector determines which frames reach the generator; the Evidence-Budget
Allocator determines how latent capacity is distributed within this fixed set.
It receives frame-level visual features and the question representation, then
produces a continuous allocation distribution and an integer token budget for
each frame.

\noindent\textbf{Instantiation.}
For the $i$-th selected frame, let
$\mathbf{h}_{i}\in\mathbb{R}^{d_v}$ denote the global frame embedding returned
by the frozen visual encoder used by the selector. We compute its compatibility
with the question as $r_i=\mathcal{A}_{\phi}(\mathbf{h}_{i},\mathbf{u}_{q})$ and
normalize the scores with temperature $\tau_a$:
\begin{equation}
\alpha_i
=\frac{\exp(r_i/\tau_a)}{\sum_{j=1}^{K}\exp(r_j/\tau_a)}.
\label{eq:allocation-distribution}
\end{equation}
The resulting $\boldsymbol{\alpha}$ describes each selected frame's relative
share of the residual latent capacity for the current question.

\noindent\textbf{Budget assignment.}
At inference, we project this continuous distribution into integer counts:
\begin{equation}
\mathbf{n}
=\mathrm{BudgetAssign}(\boldsymbol{\alpha};B,K,N_{\max}).
\label{eq:evidence-allocation}
\end{equation}
BudgetAssign first gives every selected frame one token, distributes the residual
budget according to $\boldsymbol{\alpha}$, and applies deterministic rounding
with residual correction. Hence $1\le n_i\le N_{\max}$ while the total number
of latent slots remains exactly $B$.

\subsection{Query-Aware Latent Evidence Aggregation}
\label{sec:evidence_aggregation}

The allocation $\mathbf{n}$ determines how many latent slots are assigned to
each selected frame. The aggregation head determines what these slots encode by
combining frame-specific visual content with a query-aware context shared across
all selected frames.

\noindent\textbf{Query-visual context.}
We use question-to-visual cross-attention followed by lightweight fusion to
aggregate the unpooled visual tokens into a shared query-visual context:
\begin{equation}
\mathbf{c}_{qv}
=\mathrm{Agg}_{\eta}(\mathbf{u}_{q},\mathbf{H}^{v})
\in\mathbb{R}^{d}.
\label{eq:query-visual-context}
\end{equation}
This shared context coordinates evidence formation across frames, while the
global frame embedding $\mathbf{h}_i$ preserves frame-specific content; $d$ is
the language-model hidden dimension.

\noindent\textbf{Frame-wise evidence formation.}
For position $i$ in the selected-frame sequence, we take the first $n_i$ rows
from a trainable latent basis
$\mathbf{Z}_{i}\in\mathbb{R}^{N_{\max}\times d}$, shared across examples,
and condition them on both local and shared context:
\begin{equation}
\widetilde{\mathbf{E}}_i
=\mathbf{Z}_{i,1:n_i}
+\mathbf{1}_{n_i}
\mathrm{MLP}_{\eta}
(\mathbf{W}_v\mathbf{h}_i+\mathbf{W}_c\mathbf{c}_{qv})^{\top}.
\label{eq:frame-evidence-formation}
\end{equation}
Here $\mathbf{W}_v$ and $\mathbf{W}_c$ are learned projections,
$\mathbf{1}_{n_i}$ broadcasts the conditioning vector across the allocated
slots. The basis vectors provide reusable slot identities, whereas the
conditioning term injects example-specific information. Thus,
$\widetilde{\mathbf{E}}_i\in\mathbb{R}^{n_i\times d}$ is a variable-length,
frame-specific latent bundle whose capacity is controlled by $n_i$.

\noindent\textbf{Evidence integration.}
The preliminary bundles encode frame-specific content and shared query-visual
context, but have not yet interacted with the complete explicit multimodal
input. We concatenate them in temporal order as
$\widetilde{\mathbf{E}}=\mathrm{Concat}_{i=1}^{K}
\widetilde{\mathbf{E}}_i$, yielding exactly $B$ preliminary slots. Let
$\mathbf{X}^{v}=P_{\mathrm{vis}}(\mathbf{H}^{v})$ denote the explicit visual
embeddings and
$\mathbf{X}_{0}=\mathrm{Pack}(\mathbf{X}^{q},\mathbf{X}^{v})$ denote the
explicit multimodal input. We instantiate the evidence integrator using a LoRA
adapter $\delta$ attached to $F_{\mathrm{LLM}}$. The
preliminary slots are appended to $\mathbf{X}_{0}$, and their final-layer
hidden states form the integrated latent evidence:
\begin{equation}
\mathbf{E}
=\left[
F_{\mathrm{LLM}}^{\delta}\!\left(
\mathrm{Pack}(\mathbf{X}_{0},\widetilde{\mathbf{E}})
\right)
\right]_{\mathrm{lat}}
\in\mathbb{R}^{B\times d}.
\label{eq:evidence-formation}
\end{equation}
Here $[\cdot]_{\mathrm{lat}}$ selects the hidden states at the appended latent
positions. Only $\delta$ is updated, leaving the language-model backbone
frozen. This refinement lets the preliminary slots absorb the full
multimodal context in the backbone's native hidden space. Allocation
therefore determines how many slots each frame receives, while evidence
integration determines how those slots are contextualized for the target
Video-MLLM.

\subsection{Joint Optimization}
\label{sec:joint_optimization}

We jointly train the allocator and evidence adapter from answer likelihood,
using relative likelihood to compare allocations and absolute likelihood to
supervise evidence construction.
Because the integer allocation is discrete, after the one-token floor we let
$R=B-K$ be the residual budget and draw $G$ allocation samples per example
($G=8$) for within-example comparison:
\begin{equation}
\mathbf{b}^{(g)}\sim
\mathrm{Multinomial}(R,\boldsymbol{\alpha}),
\qquad n_i^{(g)}=1+b_i^{(g)}.
\label{eq:allocation-sampling}
\end{equation}
For each sampled allocation, the evidence adapter constructs
$\mathbf{E}^{(g)}$, and the language model with frozen base parameters evaluates
the average teacher-forced answer log-likelihood
$\ell_{\mathrm{lat}}(\mathbf{n}^{(g)})$ under
$\mathrm{Pack}(\mathbf{X}_0,\mathbf{E}^{(g)})$. Let
$\bar{\ell}_{\mathrm{lat}}=G^{-1}\sum_{g=1}^{G}
\ell_{\mathrm{lat}}(\mathbf{n}^{(g)})$ and let $\sigma_{\mathrm{lat}}$ be the
corresponding standard deviation. The within-group advantage is
$A_g=(\ell_{\mathrm{lat}}(\mathbf{n}^{(g)})-
\bar{\ell}_{\mathrm{lat}})/(\sigma_{\mathrm{lat}}+\epsilon)$, where
$\epsilon>0$ stabilizes normalization.

Let $\ell_{\mathrm{exp}}$ denote the corresponding likelihood under the
explicit input $\mathbf{X}_0$ alone and define the latent gain as
$\Delta=\bar{\ell}_{\mathrm{lat}}-\mathrm{sg}(\ell_{\mathrm{exp}})$. With
$\mathrm{sg}$ denoting stop-gradient, we jointly maximize
\begin{equation}
\small
\begin{split}
\max_{\phi,\eta}\;&
\frac{1}{G}\sum_{g=1}^{G}\mathrm{sg}(A_g)
\log p_{\mathrm{Multi}}
(\mathbf{b}^{(g)};R,\boldsymbol{\alpha})\\
&+\bar{\ell}_{\mathrm{lat}}
-\lambda_{\mathrm{m}}\max(0,m-\Delta).
\end{split}
\label{eq:training-objective}
\end{equation}
Here $p_{\mathrm{Multi}}$ is the multinomial probability mass,
$\lambda_{\mathrm{m}}$ weights the margin term, and $m$ is its target gain. The
first term is a normalized score-function surrogate that updates the allocator
$\phi$ toward allocations with higher relative latent likelihood. The
likelihood and margin terms update the evidence adapter $\eta$ without
differentiating through the sampled counts. The margin penalty is active only
when $\bar{\ell}_{\mathrm{lat}}<\ell_{\mathrm{exp}}+m$ and vanishes once the
target margin is reached. The condition $N_{\max}\ge B-K+1$ keeps all sampled
counts within the per-frame latent-basis capacity. At inference, multinomial
sampling is replaced by the deterministic BudgetAssign operator in
Eq.~\eqref{eq:evidence-allocation}. Here $\eta$ comprises the latent basis,
aggregator, projections, conditioning MLP, and LoRA parameters $\delta$. Only
$(\phi,\eta)$ are updated; the base parameters of the temporal selector and
Video-MLLM remain frozen. Labels at question,
explicit-visual, and latent-evidence positions are masked, and supervision is
applied only at answer positions.

\subsection{Adaptive Evidence Invocation}
\label{sec:adaptive_evidence_invocation}

Rather than introducing a separately trained router, Adaptive Evidence
Invocation reuses the query-conditioned allocation already produced for
evidence construction. Its geometry indicates whether latent capacity is
concentrated on a few selected frames or distributed across the sequence. We
use this allocation geometry as an intrinsic invocation signal: dispersed
allocations call for integrating frame-specific evidence across the sequence,
whereas concentrated allocations retain the explicit-evidence route. The same
allocation therefore controls both evidence capacity and adaptive route
selection, without an auxiliary routing module or a separate routing forward
pass. By inserting latent evidence only when invoked, this design also reduces
the average decoder-side token overhead relative to always-on insertion.

\noindent\textbf{Allocation dispersion.}
Every selected frame receives one base token to guarantee coverage, but this
floor does not reflect relative evidence demand. We therefore remove it and
normalize the residual allocation as
$\widetilde{p}_i=(n_i-1)/(B-K)$. We measure how broadly the residual capacity is
distributed using a normalized $\ell_2$-based dispersion:
\begin{equation}
s_{\mathrm{disp}}(\mathbf{n})
=\frac{1-\sum_{i=1}^{K}\widetilde{p}_i^2}{1-1/K}.
\label{eq:allocation-dispersion}
\end{equation}
The numerator is the probability that two residual-token draws are assigned to
different selected frames, while the denominator maps a uniform allocation to
one. Thus $s_{\mathrm{disp}}\in[0,1]$: zero places all residual capacity on one
frame, whereas one distributes it uniformly across the $K$ frames.

\noindent\textbf{Adaptive route selection.}
We select the route of each query using the fixed boundary
$s_{\mathrm{disp}}=1/2$:
\begin{equation}
a=\mathbb{I}\!\left[s_{\mathrm{disp}}(\mathbf{n})>\frac{1}{2}\right].
\label{eq:adaptive-invocation}
\end{equation}
This boundary has a direct effective-support interpretation. Defining
$N_{\mathrm{eff}}=(\sum_i\widetilde{p}_i^2)^{-1}$, the latent route is selected
when $N_{\mathrm{eff}}>2K/(K+1)$, a threshold close to two frames for our frame
budgets. It therefore invokes cross-frame integration when residual capacity is
substantively shared across multiple frames, while retaining the explicit route
for concentrated allocations.
Recall that $\mathbf{X}_{0}=\mathrm{Pack}(\mathbf{X}^{q},\mathbf{X}^{v})$
denotes the explicit input. When $a=0$, the Explicit Evidence Route skips the
integration step in Eq.~\eqref{eq:evidence-formation} and sends $\mathbf{X}_{0}$
directly to $F_{\mathrm{LLM}}$. When $a=1$, the Latent Evidence Route first
constructs $\mathbf{E}$ using Eq.~\eqref{eq:evidence-formation} and forms
$\mathbf{X}_{1}=\mathrm{Pack}(\mathbf{X}_{0},\mathbf{E})$. The integration
adapter $\delta$ is disabled during answer generation. Both routes retain the
explicit evidence and use a single autoregressive answer-generation pass
through the frozen backbone.
Adaptive Evidence Invocation is applied only at inference; training uses the
paired explicit and sampled-latent likelihoods described above.

\section{Experiments}

We conduct experiments to answer the following research questions:
\begin{itemize}[leftmargin=*,itemsep=0.1em,topsep=0.2em]
    \item (\textbf{RQ1}) How effective is \ours{} for long-video understanding?
    \item (\textbf{RQ2}) How efficiently does \ours{} use compact visual evidence?
    \item (\textbf{RQ3}) Does learned allocation reflect query-specific evidence structure?
    \item (\textbf{RQ4}) Does Adaptive Evidence Invocation exploit route complementarity?
\end{itemize}

\subsection{Experimental Settings}
\noindent\textbf{Benchmarks and metrics.} We report accuracy (\%) on
LongVideoBench~\cite{wu2024longvideobench},
MLVU~\cite{zhou2025mlvu} (\textit{M-Avg} development split),
Video-MME~\cite{fu2025videomme} (overall score without subtitles), and
LVBench~\cite{wang2025lvbench}.

\noindent\textbf{Implementation.} We freeze both Video-MLLM backbones and the
pretrained temporal selector, including its CLIP encoder. Trainable components
include the allocation head, latent evidence basis, context aggregator,
projection layers, conditional MLP, and LoRA parameters. Unless specified
otherwise, we use $K$ selected frames and $B=32$ latent tokens. The selector
uses the public TSPO checkpoint, and the adapter is trained on 8,000 samples
from LLaVA-Video-178K using eight NVIDIA A800 GPUs.

\subsection{Main Results}

\paragraph{[For RQ1] \ours{} consistently improves long-video understanding.}
Table~\ref{tab:video_overall_comparison} compares \ours{} with recent
Video-MLLMs and matched-frame base models. The matched-frame cells keep the
backbone and frame budget fixed and report Base/\ours{} accuracy. Under this
protocol, \ours{} improves the four-benchmark LLaVA-Video average by
$+5.2/+3.4$ points at 8/16 frames. On Qwen2.5-VL, its largest gains are
$+10.1/+8.4$ points on LVBench. At 16 frames, \ours{} with LLaVA-Video
outperforms several frame-selection methods, such as TSPO and
DIG, on all shared benchmarks and matches or surpasses Vamba's 1024-frame
results. It also outperforms representative multi-turn selectors, such as
Video-MTR and LongVT, on their reported metrics despite their larger budgets,
and several reasoning models, such as Video-R1, Conan, and VTI-CoT, on every
available comparison. Across these families, \ours{} uses selected evidence
more effectively, outperforming methods with larger frame budgets or iterative
retrieval.

\providecommand{\scoregain}[1]{\textcolor{mygreen}{\scriptsize\,(\ensuremath{\uparrow}#1)}}

\begin{table}[t]
\centering
\small
\setlength{\tabcolsep}{3pt}
\renewcommand{\arraystretch}{1.0}

\resizebox{\linewidth}{!}{%
\begin{tabular}{lcccccc}
\toprule
\textbf{Model}
& \textbf{Size}
& \textbf{\#Frames}
& \textbf{Video-MME}
& \textbf{MLVU}
& \textbf{LongVideoBench}
& \textbf{LVBench} \\
\midrule

\rowcolor{tablerowcolor1}
\multicolumn{7}{l}{$\blacktriangledown$ \emph{Proprietary Models}} \\
GPT-4V
& -- & 1fps
& 60.7 & -- & -- & -- \\
GPT-4o
& -- & 1fps
& 77.2 & 66.2 & 66.7 & 34.7 \\

\midrule
\rowcolor{tablerowcolor}
\multicolumn{7}{l}{$\blacktriangledown$ \emph{Open-Source Video MLLMs}} \\
Video-LLaVA~\cite{lin2024video}
& 7B & 8
& 40.4 & 47.3 & 39.1 & -- \\
LongVA~\cite{zhang2024long}
& 7B & 128
& 54.3 & 56.3 & -- & -- \\
LLaVA-OneVision~\cite{li2024llava}
& 7B & 32
& 58.2 & 64.7 & -- & -- \\
VideoChat-T~\cite{zeng2025timesuite}
& 7B & 12
& 46.3 & -- & -- & -- \\
Video-XL~\cite{shu2025video}
& 7B & 256
& 55.5 & 64.9 & 50.7 & -- \\
Video-XL-Pro~\cite{liu2025video}
& 7B & 240
& 60.0 & 70.6 & 56.7 & -- \\
Vamba~\cite{ren2025vamba}
& 10B & 1024
& 57.8 & 65.9 & 55.9 & 42.1 \\
LongVILA~\cite{chen2025longvila}
& 7B & 256
& 60.1 & -- & 57.1 & -- \\
LongVU~\cite{shen2024longvu}
& 7B & 1fps
& 60.6 & 65.4 & -- & -- \\
TSPO~\cite{tang2026tspo}
& 7B & 16
& 55.3 & 57.7 & 54.8 & 38.6 \\
DIG~\cite{li2026divide} & 7B & 16 & -- & 66.21 & 58.86 & -- \\
EFS~\cite{chen2026efs} & 7B & 16 & 60.0 & 66.0 & 60.5 & -- \\
Video Panels~\cite{doorenbos2026videopanels}
& 7B & 32
& 63.9 & 64.9 & -- & -- \\
ReaSon~\cite{zhou2026reason}
& 7B & 32
& 57.9 & -- & -- & -- \\
LongVT~\cite{yang2026longvt}& 7B & 64/512/768 & -- & --& --& 41.3 \\
VideoBrain~\cite{zou2026videobrain} & 8B & >20 &-- & --& 53.3 & 41.3 \\
\midrule
\rowcolor{tablerowcolor}
\multicolumn{7}{l}{$\blacktriangledown$ \emph{Open-Source Reasoning Video MLLMs}} \\
Video-MTR~\cite{xie2026videomtr}
& 7B & 32
& 59.0 & -- & -- & -- \\
Video-R1~\cite{feng2025videor1}
& 7B & 32
& 59.3 & -- & -- & -- \\
Conan~\cite{ouyang2025conan}
& 7B & 32
& 60.5 & 63.4 & 56.6 & 39.2 \\
VTI-CoT~\cite{zhang2026vticot} & 7B & 32
& 59.6 & -- & 55.0 & 40.5 \\

\midrule
\rowcolor{tablerowcolor}
\multicolumn{7}{l}{$\blacktriangledown$ \emph{Base Models and \ours{} under matched frame budgets}} \\
LLaVA-Video (Base)
& 7B & 8
& 56.3 & 60.5 & 54.9 & 35.5 \\
\rowcolor{gray!18}
\quad\textbf{+\ours}
& \textbf{7B} & \textbf{8}
& \textbf{58.7}\scoregain{2.4} & \textbf{69.4}\scoregain{8.9} & \textbf{58.4}\scoregain{3.5} & \textbf{41.4}\scoregain{5.9} \\
LLaVA-Video (Base)
& 7B & 16
& 60.5 & 64.5 & 57.5 & 37.9 \\
\rowcolor{gray!18}
\quad\textbf{+\ours}
& \textbf{7B} & \textbf{16}
& \textbf{61.4}\scoregain{0.9} & \textbf{70.8}\scoregain{6.3} & \textbf{59.5}\scoregain{2.0} & \textbf{42.1}\scoregain{4.2} \\
Qwen2.5-VL (Base)
& 7B & 8
& 52.9 & 53.3 & 52.3 & 33.1 \\
\rowcolor{gray!18}
\quad\textbf{+\ours}
& \textbf{7B} & \textbf{8}
& \textbf{55.9}\scoregain{3.0} & \textbf{58.2}\scoregain{4.9} & \textbf{53.4}\scoregain{1.1} & \textbf{43.2}\scoregain{10.1} \\
Qwen2.5-VL (Base)
& 7B & 16
& 56.7 & 56.7 & 55.2 & 35.2 \\
\rowcolor{gray!18}
\quad\textbf{+\ours}
& \textbf{7B} & \textbf{16}
& \textbf{58.9}\scoregain{2.2} & \textbf{63.3}\scoregain{6.6} & \textbf{55.7}\scoregain{0.5} & \textbf{43.6}\scoregain{8.4} \\

\bottomrule
\end{tabular}%
}
\caption{Comparison with recent Video-MLLMs.}
\label{tab:video_overall_comparison}
\end{table}

\paragraph{[For RQ2] \ours{} improves accuracy without expanding explicit visual context.}

\begin{figure}[t]
    \centering
    \includegraphics[width=0.98\textwidth]{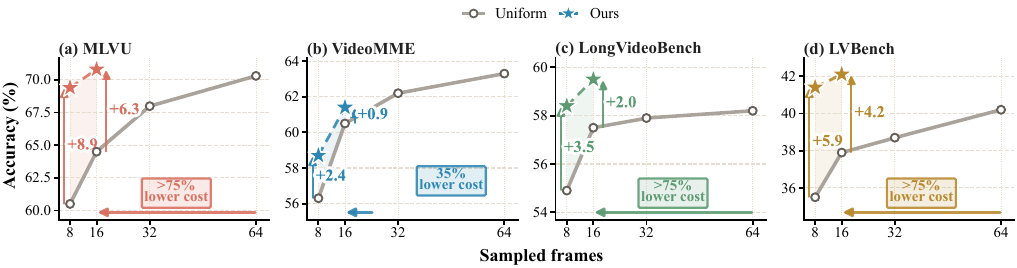}
    \caption{Frame efficiency against uniform sampling. Vertical arrows denote
    same-frame gains; horizontal arrows denote frame reductions at matched accuracy.}
    \label{fig:frame_efficiency}
\end{figure}

Figure~\ref{fig:frame_efficiency} tests whether a compact visual context
augmented by \ours{} can rival larger explicit contexts.
At 8/16 frames, \ours{} improves MLVU by $8.9/6.3$, Video-MME by $2.4/0.9$,
LongVideoBench by $3.5/2.0$, and LVBench by $5.9/4.2$ points. Its 16-frame
setting also outperforms 64-frame uniform sampling on MLVU, LongVideoBench, and
LVBench and matches roughly 24 uniform frames on Video-MME.
Thus, \ours{} extracts more answer-relevant value from compact visual input
instead of relying on additional frames.

\begin{figure}[t]
    \centering
    \begin{minipage}[t]{0.48\textwidth}
        \vspace{0pt}
        \centering
        \includegraphics[width=\linewidth]{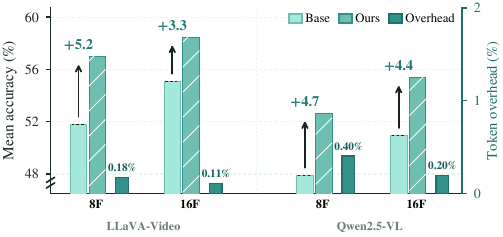}
        \caption{Token-cost analysis under matched frame budgets.}
        \label{fig:token_cost_gain}
    \end{minipage}
    \hfill
    \begin{minipage}[t]{0.48\textwidth}
        \vspace{0pt}
        \centering
        \includegraphics[width=\linewidth]{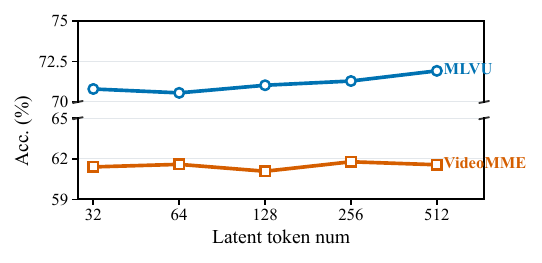}
        \caption{Latent-token budget sensitivity.}
        \label{fig:latent_budget_sensitivity}
    \end{minipage}
\end{figure}

Adaptive Evidence Invocation preserves this compactness by inserting latent
evidence only when invoked. For visual-token cost $N_{\mathrm{vis}}$, latent
budget $B$, and empirical activation rate $\rho$, its decoder-side overhead is
$\rho B/N_{\mathrm{vis}}\times100\%$.
Figure~\ref{fig:token_cost_gain} reports only $0.11\%$--$0.40\%$ decoder-side
video-token overhead for mean gains of $+3.3$--$+5.2$ points; full accounting
is provided in the appendix.

Figure~\ref{fig:latent_budget_sensitivity} shows limited returns from increasing
$B$ from $32$ to $512$: MLVU rises from $70.8$ to $71.9$, while Video-MME
improves by at most $+0.4$. We therefore use $B=32$, retaining most of the
observed gain without lengthening the latent sequence.

% Derived from tables/inference_latency.csv.
\begin{wraptable}{r}{0.44\textwidth}
\vspace{-2em}
\centering
\small
\setlength{\tabcolsep}{2.5pt}
\renewcommand{\arraystretch}{1.08}
\resizebox{\linewidth}{!}{%
\begin{tabular}{@{}lcccc@{}}
\toprule
\textbf{Setting} & \textbf{Video-MME} & \textbf{MLVU} &
\textbf{LongVideoBench} & \textbf{LVBench} \\
\midrule
Explicit & 1.403 & 7.453 & 1.472 & 0.831 \\
\textbf{\ours{}} & 1.468 & 7.481 & 1.602 & 0.834 \\
\bottomrule
\end{tabular}%
}
\caption{Average LLaVA-Video inference time (s/example, $\downarrow$) at 16 frames.}
% \vspace{-3em}
\label{tab:inference_efficiency}
\end{wraptable}

Table~\ref{tab:inference_efficiency} reports end-to-end inference latency,
including allocation, aggregation, and adapter computation. At 16 frames,
average latency increases from $2.790$ to $2.847$ seconds per example
($+2.0\%$), with an
absolute increase of at most $0.130$ seconds across benchmarks.

\paragraph{[For RQ3] \ours{} adapts latent capacity to localized and distributed evidence.}
\begin{figure}[t]
    \centering
    \includegraphics[width=0.98\textwidth]{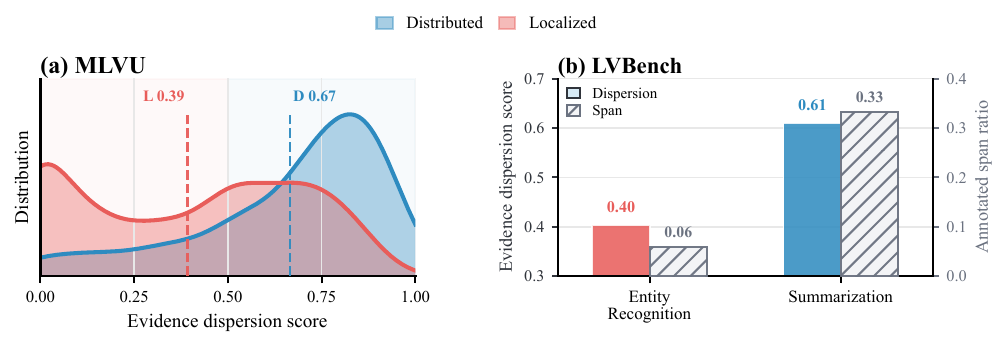}
    \caption{Allocation dispersion follows evidence scope on MLVU and LVBench.
    Higher values indicate residual latent capacity spread across more frames.}
    \label{fig:rq3_allocation_density}
\end{figure}

An evidence allocator should concentrate latent capacity when a query hinges on
localized cues and spread it when answering requires evidence across frames.
Figure~\ref{fig:rq3_allocation_density} tests this expectation using benchmark
groups defined independently of the learned allocation. After removing the
mandatory base token, $s_{\mathrm{disp}}$ measures how broadly residual capacity
is spread across the selected frames. On MLVU, holistic tasks exhibit higher
dispersion than single-detail tasks ($0.67$ vs.\ $0.39$). On LVBench,
summarization covers longer annotated spans than entity recognition
($0.33$ vs.\ $0.06$ of the video) and likewise exhibits higher dispersion
($0.61$ vs.\ $0.40$). This localized-to-distributed shift at both task and
instance levels shows that allocation geometry tracks the evidence scope of a
query. The resulting geometry supplies Adaptive Evidence Invocation with an
intrinsic signal for cross-frame integration; RQ4 tests its effect on answers.

\paragraph{[For RQ4] Adaptive Evidence Invocation converts route complementarity into higher accuracy.}
\begin{wrapfigure}{l}{0.48\textwidth}
    \centering
    \includegraphics[width=\linewidth]{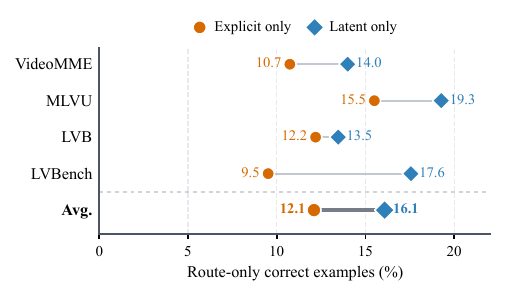}
    \caption{Complementary route outcomes on 7,703 paired 8-frame examples
    pooled across all four benchmarks; each pair uses identical selected frames.}
    \label{fig:route_counterfactual}
\end{wrapfigure}

Adaptive routing is motivated by route complementarity.
Figure~\ref{fig:route_counterfactual} pools all four benchmarks and compares the
Explicit and Latent Evidence Routes under identical selected frames. Among the
resulting 7,703 paired 8-frame examples, only the latent route is correct on
$16.1\%$, while only the explicit route is correct on $12.1\%$. Neither route
dominates, motivating per-query route selection.

Table~\ref{tab:ours_ablation} then evaluates the allocation-derived signal.
The Full, Random, and Adaptive policies share the learned allocator and
aggregation head, isolating the invocation rule. Adaptive reaches $52.7/55.4$,
outperforming Full ($52.3/54.2$) and Random ($50.3/52.9$) at both frame budgets.
The same distribution can therefore govern both latent capacity and evidence
use.

\begin{table}[t]
\centering
{\small
\setlength{\tabcolsep}{2.5pt}
\renewcommand{\arraystretch}{1.12}
\resizebox{\linewidth}{!}{%
\begin{tabular}{@{}lcccccccc@{}}
\toprule
\textbf{Variant} & \textbf{Selector} & \textbf{Latent} & \textbf{Invocation}
& \textbf{Video-MME} & \textbf{MLVU} & \textbf{LongVideoBench} & \textbf{LVBench} & \textbf{Avg.} \\
\midrule
Vanilla & Uniform & \xmark & --
& 52.9/56.7 & 53.3/56.7 & 52.3/55.2 & 33.1/35.2 & 47.9/51.0 \\
Evidence localization & Frozen & \xmark & --
& 52.5/55.3 & 53.8/57.7 & 51.6/54.8 & 35.1/38.6 & 48.3/51.6 \\
Uniform allocation & Frozen & \cmark & Always
& 54.4/56.2 & 56.7/59.8 & 52.3/52.1 & 38.9/40.5 & 50.6/52.2 \\
Full invocation & Frozen & \cmark & Always
& 55.8/58.2 & 57.6/62.6 & 52.7/52.8 & 43.2/43.2 & 52.3/54.2 \\
Random invocation & Frozen & \cmark & Random ($p=0.5$)
& 54.2/56.8 & 55.7/60.2 & 52.2/53.8 & 39.2/40.9 & 50.3/52.9 \\
\textbf{\ours} & Frozen & \cmark & Adaptive
& \textbf{55.9/58.9} & \textbf{58.2/63.3} & \textbf{53.4/55.7} & \textbf{43.2/43.6} & \textbf{52.7/55.4} \\
\bottomrule
\end{tabular}%
}
}
\caption{Ablation study. Each result reports 8-frame/16-frame
accuracy (\%); Avg. averages the four benchmarks at the corresponding frame
budget.}
\label{tab:ours_ablation}
\end{table}

The remaining variants trace the evidence-use pipeline. Frozen evidence
localization contributes $+0.4/+0.6$ points over uniform sampling. Adding latent
aggregation with a uniform budget reaches $50.6/52.2$, and learned allocation
adds $+1.7/+2.0$ under Full invocation. Adaptive invocation completes the
progression, yielding overall gains of $+4.8/+4.4$ points at 8/16 frames. This
separates the contribution of post-selection evidence use from selection alone.

\section{Conclusion}

We presented \ours{}, a distribution-guided latent evidence aggregation
framework for compact long-video understanding. \ours{} constructs query-conditioned
latent evidence from selected frames and invokes it according to the
resulting allocation dispersion. Across four long-video benchmarks and two
Video-MLLM backbones, \ours{} improves matched-frame baselines with a compact
latent-token budget. The allocation and ablation analyses further show
task-aware allocation patterns and average gains from Adaptive Evidence
Invocation over Full invocation and Random invocation at both frame budgets.
Together, these results establish the latent evidence interface as a
complementary post-selection stage between frame selection and answer generation.

\bibliographystyle{assets/plainnat}
\bibliography{reference}

\begin{thebibliography}{62}
\providecommand{\natexlab}[1]{#1}
\providecommand{\url}[1]{\texttt{#1}}
\expandafter\ifx\csname urlstyle\endcsname\relax
  \providecommand{\doi}[1]{doi: #1}\else
  \providecommand{\doi}{doi: \begingroup \urlstyle{rm}\Url}\fi

\bibitem[Ahn et~al.(2026)Ahn, Han, Kim, Lee, and Choi]{ahn2026evident}
Geo Ahn, Jiwook Han, Youngrae Kim, Joonseok Lee, and Jinwoo Choi.
\newblock {EVIDENT}: Routing mllm adaptation through entity-grounded visual evidence for cross-domain video temporal grounding.
\newblock \emph{arXiv preprint arXiv:2605.26104}, 2026.

\bibitem[Asadi et~al.(2026)Asadi, O'Sullivan, Cao, Nedaee, Rajabalifardi, Li, Adeli, and Ashley]{asadi2026mirage}
Mohammad Asadi, Jack~W. O'Sullivan, Fang Cao, Tahoura Nedaee, Kamyar Rajabalifardi, Fei-Fei Li, Ehsan Adeli, and Euan Ashley.
\newblock {MIRAGE}: The illusion of visual understanding.
\newblock \emph{arXiv preprint arXiv:2603.21687}, 2026.

\bibitem[Azadani et~al.(2026)Azadani, Wang, Zhu, Chen, Ganai, Sedwards, Pavone, and Czarnecki]{azadani2026vistaqa}
Mozhgan~Nasr Azadani, Yimu Wang, Yongpeng Zhu, Lihong Chen, Milan Ganai, Sean Sedwards, Marco Pavone, and Krzysztof Czarnecki.
\newblock {VISTAQA}: Benchmarking joint visual question answering and pixel-level evidence.
\newblock \emph{arXiv preprint arXiv:2605.20676}, 2026.

\bibitem[Ben-Levi et~al.(2026)Ben-Levi, Goldfeder, Zhao, Lapid, LeVi, Roush, Shwartz-Ziv, and Lipson]{benlevi2026mirageprobes}
Daniel Ben-Levi, Judah Goldfeder, Weiliang Zhao, Raz Lapid, Amit LeVi, Allen~G. Roush, Ravid Shwartz-Ziv, and Hod Lipson.
\newblock Mirage probes: How vision models fake visual understanding.
\newblock \emph{arXiv preprint arXiv:2606.13870}, 2026.

\bibitem[Chen et~al.(2023)Chen, Zhang, Zeng, Zhang, Zhu, and Zhao]{chen2023shikra}
Keqin Chen, Zhao Zhang, Weili Zeng, Richong Zhang, Feng Zhu, and Rui Zhao.
\newblock Shikra: Unleashing multimodal llm's referential dialogue magic.
\newblock \emph{arXiv preprint arXiv:2306.15195}, 2023.

\bibitem[Chen et~al.(2024{\natexlab{a}})Chen, Wei, Li, Dong, Zhang, Zang, Chen, Duan, Lin, Tang, et~al.]{chen2024sharegpt4video}
Lin Chen, Xilin Wei, Jinsong Li, Xiaoyi Dong, Pan Zhang, Yuhang Zang, Zehui Chen, Haodong Duan, Bin Lin, Zhenyu Tang, et~al.
\newblock Sharegpt4video: Improving video understanding and generation with better captions.
\newblock \emph{Advances in Neural Information Processing Systems}, 37:\penalty0 19472--19495, 2024{\natexlab{a}}.

\bibitem[Chen et~al.(2026)Chen, Luo, Zeng, Lin, Xie, Chao, Ji, and Zheng]{chen2026efs}
Wang Chen, Yongdong Luo, Yuhui Zeng, Luojun Lin, Tianyu Xie, Fei Chao, Rongrong Ji, and Xiawu Zheng.
\newblock Event-anchored frame selection for effective long-video understanding.
\newblock In \emph{Proceedings of the IEEE/CVF Conference on Computer Vision and Pattern Recognition}, 2026.

\bibitem[Chen et~al.(2025)Chen, Xue, Li, Hu, Zhu, Li, Fang, Tang, Yang, Liu, et~al.]{chen2025longvila}
Yukang Chen, Fuzhao Xue, Dacheng Li, Qinghao Hu, Ligeng Zhu, Xiuyu Li, Yunhao Fang, Haotian Tang, Shang Yang, Zhijian Liu, et~al.
\newblock Longvila: Scaling long-context visual language models for long videos.
\newblock In \emph{International Conference on Learning Representations}, volume 2025, pages 18227--18246, 2025.

\bibitem[Chen et~al.(2024{\natexlab{b}})Chen, Wang, Cao, Liu, Gao, Cui, Zhu, Ye, Tian, Liu, et~al.]{chen2024expanding}
Zhe Chen, Weiyun Wang, Yue Cao, Yangzhou Liu, Zhangwei Gao, Erfei Cui, Jinguo Zhu, Shenglong Ye, Hao Tian, Zhaoyang Liu, et~al.
\newblock Expanding performance boundaries of open-source multimodal models with model, data, and test-time scaling.
\newblock \emph{arXiv preprint arXiv:2412.05271}, 2024{\natexlab{b}}.

\bibitem[Cheng et~al.(2024)Cheng, Leng, Zhang, Xin, Li, Chen, Zhu, Zhang, Luo, Zhao, et~al.]{cheng2024videollama}
Zesen Cheng, Sicong Leng, Hang Zhang, Yifei Xin, Xin Li, Guanzheng Chen, Yongxin Zhu, Wenqi Zhang, Ziyang Luo, Deli Zhao, et~al.
\newblock Videollama 2: Advancing spatial-temporal modeling and audio understanding in video-llms.
\newblock \emph{arXiv preprint arXiv:2406.07476}, 2024.

\bibitem[Cui et~al.(2026)Cui, Long, Zhang, Zhang, Su, Gan, Zhao, and Ren]{cui2026ris}
Jin Cui, Xinyue Long, Xunyong Zhang, Yadong Zhang, Chuanchang Su, Jingye Gan, Boran Zhao, and Pengju Ren.
\newblock Retrieve, integrate, and synthesize: Spatial-semantic grounded latent visual reasoning.
\newblock \emph{arXiv preprint arXiv:2605.07106}, 2026.

\bibitem[Doorenbos et~al.(2026)Doorenbos, Spurio, and Gall]{doorenbos2026videopanels}
Lars Doorenbos, Federico Spurio, and Juergen Gall.
\newblock Video panels for long video understanding.
\newblock In \emph{Proceedings of the IEEE/CVF Conference on Computer Vision and Pattern Recognition}, 2026.

\bibitem[Du et~al.(2026)Du, Liu, Yu, Ding, Wu, Wang, Nie, Liu, Chen, Song, and Li]{du2026medhorizon}
Bodong Du, Bowen Liu, Yang Yu, Xinpeng Ding, Zhiheng Wu, Shuning Wang, Shuo Nie, Naiming Liu, Qifeng Chen, Yangqiu Song, and Xiaomeng Li.
\newblock Medhorizon: Towards long-context medical video understanding in the wild, 2026.
\newblock \url{https://arxiv.org/abs/2605.06537}.

\bibitem[Favero et~al.(2024)Favero, Zancato, Trager, Choudhary, Perera, Achille, Swaminathan, and Soatto]{favero2024multimodal}
Alessandro Favero, Luca Zancato, Matthew Trager, Siddharth Choudhary, Pramuditha Perera, Alessandro Achille, Ashwin Swaminathan, and Stefano Soatto.
\newblock Multi-modal hallucination control by visual information grounding.
\newblock In \emph{Proceedings of the IEEE/CVF Conference on Computer Vision and Pattern Recognition}, pages 14303--14312, 2024.
\newblock \doi{10.1109/CVPR52733.2024.01356}.

\bibitem[Feng et~al.(2025)Feng, Gong, Li, Guo, Wang, Peng, Wu, Zhang, Wang, and Yue]{feng2025videor1}
Kaituo Feng, Kaixiong Gong, Bohao Li, Zonghao Guo, Yibing Wang, Tianshuo Peng, Junfei Wu, Xiaoying Zhang, Benyou Wang, and Xiangyu Yue.
\newblock Video-{R1}: Reinforcing video reasoning in {MLLM}s.
\newblock In \emph{Advances in Neural Information Processing Systems}, volume~38, pages 99114--99137. Curran Associates, Inc., 2025.
\newblock \url{https://proceedings.neurips.cc/paper_files/paper/2025/file/8eb3976840e08b80dda9667562574246-Paper-Conference.pdf}.

\bibitem[Fu et~al.(2025)Fu, Dai, Luo, Li, Ren, Zhang, Wang, Zhou, Shen, Zhang, et~al.]{fu2025videomme}
Chaoyou Fu, Yuhan Dai, Yongdong Luo, Lei Li, Shuhuai Ren, Renrui Zhang, Zihan Wang, Chenyu Zhou, Yunhang Shen, Mengdan Zhang, et~al.
\newblock Video-{MME}: The first-ever comprehensive evaluation benchmark of multi-modal {LLM}s in video analysis.
\newblock In \emph{Proceedings of the IEEE/CVF Conference on Computer Vision and Pattern Recognition}, pages 24108--24118, 2025.

\bibitem[Gong et~al.(2026)Gong, Wu, Yuan, Hu, Zhang, Zhou, Chen, Niu, Wang, Li, Wang, Qi, Ji, and Yang]{gong2026pixeleyes}
Dengxian Gong, Yuanzheng Wu, Haobo Yuan, Zhengdong Hu, Tao Zhang, Yikang Zhou, Shihao Chen, Quanzhu Niu, Kai Wang, Jason Li, Haochen Wang, Lu~Qi, Shunping Ji, and Ming-Hsuan Yang.
\newblock {PixelEyes}: Decoupling perception and reasoning for pinpoint visual evidence seeking.
\newblock \emph{arXiv preprint arXiv:2607.00115}, 2026.

\bibitem[Jia et~al.(2026)Jia, Liu, Yang, Yan, Zou, and Hu]{jia2026decoding}
Sihang Jia, Shuliang Liu, Songbo Yang, Yibo Yan, Xin Zou, and Xuming Hu.
\newblock Decoding by perturbation: Mitigating mllm hallucinations via dynamic textual perturbation.
\newblock \emph{arXiv preprint arXiv:2604.12424}, 2026.

\bibitem[Lai et~al.(2024)Lai, Tian, Chen, Li, Yuan, Liu, and Jia]{lai2023lisa}
Xin Lai, Zhuotao Tian, Yukang Chen, Yanwei Li, Yuhui Yuan, Shu Liu, and Jiaya Jia.
\newblock {LISA}: Reasoning segmentation via large language model.
\newblock In \emph{Proceedings of the IEEE/CVF Conference on Computer Vision and Pattern Recognition}, pages 9579--9589, 2024.

\bibitem[Li et~al.(2024{\natexlab{a}})Li, Zhang, Guo, Zhang, Li, Zhang, Zhang, Zhang, Li, Liu, et~al.]{li2024llava}
Bo~Li, Yuanhan Zhang, Dong Guo, Renrui Zhang, Feng Li, Hao Zhang, Kaichen Zhang, Peiyuan Zhang, Yanwei Li, Ziwei Liu, et~al.
\newblock Llava-onevision: Easy visual task transfer.
\newblock \emph{arXiv preprint arXiv:2408.03326}, 2024{\natexlab{a}}.

\bibitem[Li et~al.(2023)Li, Wong, Zhang, Usuyama, Liu, Yang, Naumann, Poon, and Gao]{li2023llavamed}
Chunyuan Li, Cliff Wong, Sheng Zhang, Naoto Usuyama, Haotian Liu, Jianwei Yang, Tristan Naumann, Hoifung Poon, and Jianfeng Gao.
\newblock Llava-med: Training a large language-and-vision assistant for biomedicine in one day, 2023.
\newblock \url{https://arxiv.org/abs/2306.00890}.

\bibitem[Li et~al.(2026{\natexlab{a}})Li, Li, Li, and Lu]{li2026divide}
Jialuo Li, Bin Li, Jiahao Li, and Yan Lu.
\newblock Divide, then ground: Adapting frame selection to query types for long-form video understanding.
\newblock In \emph{Proceedings of the IEEE/CVF Conference on Computer Vision and Pattern Recognition}, pages 11369--11380, 2026{\natexlab{a}}.

\bibitem[Li et~al.(2026{\natexlab{b}})Li, Zheng, Shen, Huang, Lu, Ni, and Guan]{li2026keeping}
Jiaqi Li, Shuntian Zheng, Yixian Shen, Jia-Hong Huang, Xiaoman Lu, Minzhe Ni, and Yu~Guan.
\newblock Keeping the evidence chain: Semantic evidence allocation for training-free token pruning in video temporal grounding.
\newblock \emph{arXiv preprint arXiv:2603.05663}, 2026{\natexlab{b}}.

\bibitem[Li et~al.(2024{\natexlab{b}})Li, Wang, He, Li, Wang, Liu, Wang, Xu, Chen, Luo, et~al.]{li2024mvbench}
Kunchang Li, Yali Wang, Yinan He, Yizhuo Li, Yi~Wang, Yi~Liu, Zun Wang, Jilan Xu, Guo Chen, Ping Luo, et~al.
\newblock Mvbench: A comprehensive multi-modal video understanding benchmark.
\newblock In \emph{Proceedings of the IEEE/CVF Conference on Computer Vision and Pattern Recognition}, pages 22195--22206, 2024{\natexlab{b}}.

\bibitem[Li et~al.(2024{\natexlab{c}})Li, Wang, and Jia]{li2024llama}
Yanwei Li, Chengyao Wang, and Jiaya Jia.
\newblock Llama-vid: An image is worth 2 tokens in large language models.
\newblock In \emph{European Conference on Computer Vision}, pages 323--340. Springer, 2024{\natexlab{c}}.

\bibitem[Li et~al.(2024{\natexlab{d}})Li, Xu, Zhang, Song, Cai, Qi, Zhou, Pan, Li, Vu, Huang, and Wang]{li2024groundinggpt}
Zhaowei Li, Qi~Xu, Dong Zhang, Hang Song, Yiqing Cai, Qi~Qi, Ran Zhou, Junting Pan, Zefeng Li, Van~Tu Vu, Zhida Huang, and Tao Wang.
\newblock {GroundingGPT}: Language enhanced multi-modal grounding model.
\newblock In \emph{Proceedings of the 62nd Annual Meeting of the Association for Computational Linguistics (Volume 1: Long Papers)}, pages 6657--6678, 2024{\natexlab{d}}.

\bibitem[Lin et~al.(2024)Lin, Ye, Zhu, Cui, Ning, Jin, and Yuan]{lin2024video}
Bin Lin, Yang Ye, Bin Zhu, Jiaxi Cui, Munan Ning, Peng Jin, and Li~Yuan.
\newblock Video-llava: Learning united visual representation by alignment before projection.
\newblock In \emph{Proceedings of the 2024 conference on empirical methods in natural language processing}, pages 5971--5984, 2024.

\bibitem[Liu et~al.(2026{\natexlab{a}})Liu, Yang, Song, Tang, Gao, Chen, Song, Chen, and Li]{liu2026dividethendiagnose}
Bowen Liu, Li~Yang, Shanshan Song, Mingyu Tang, Zhifang Gao, Qifeng Chen, Yangqiu Song, Huimin Chen, and Xiaomeng Li.
\newblock Divide-then-diagnose: Weaving clinician-inspired contexts for ultra-long capsule endoscopy videos, 2026{\natexlab{a}}.
\newblock \url{https://arxiv.org/abs/2604.21814}.

\bibitem[Liu et~al.(2025)Liu, Shu, Liu, Li, Tian, and Zhao]{liu2025video}
Xiangrui Liu, Yan Shu, Zheng Liu, Ao~Li, Yang Tian, and Bo~Zhao.
\newblock Video-xl-pro: Reconstructive token compression for extremely long video understanding.
\newblock \emph{arXiv preprint arXiv:2503.18478}, 2025.

\bibitem[Liu et~al.(2026{\natexlab{b}})Liu, Zhu, Zhao, Wang, Li, Li, Cao, Sun, Zhang, Yang, Zhong, and Yang]{liu2026moment}
Xiaolin Liu, Yilun Zhu, Xiangyu Zhao, Xuehui Wang, Yan Li, Xin Li, Haoyu Cao, Xing Sun, Shaofeng Zhang, Xu~Yang, Zhihang Zhong, and Xue Yang.
\newblock Moment-video: Diagnosing temporal fidelity of video mllms on momentary visual events.
\newblock \emph{arXiv preprint arXiv:2606.02522}, 2026{\natexlab{b}}.

\bibitem[Ma et~al.(2026)Ma, Li, Wang, Wang, Kong, Zeng, Xiao, Yang, Zhang, Wang, and He]{ma2026citevqa}
Dongsheng Ma, Jiayu Li, Zhengren Wang, Yijie Wang, Jiahao Kong, Weijun Zeng, Jutao Xiao, Jie Yang, Wentao Zhang, Bin Wang, and Conghui He.
\newblock {CiteVQA}: Benchmarking evidence attribution for trustworthy document intelligence.
\newblock \emph{arXiv preprint arXiv:2605.12882}, 2026.

\bibitem[Moor et~al.(2023)Moor, Huang, Wu, Yasunaga, Zakka, Dalmia, Reis, Rajpurkar, and Leskovec]{moor2023medflamingo}
Michael Moor, Qian Huang, Shirley Wu, Michihiro Yasunaga, Cyril Zakka, Yash Dalmia, Eduardo~Pontes Reis, Pranav Rajpurkar, and Jure Leskovec.
\newblock Med-flamingo: A multimodal medical few-shot learner, 2023.
\newblock \url{https://arxiv.org/abs/2307.15189}.

\bibitem[Morini et~al.(2026)Morini, Sarto, Cornia, and Baraldi]{morini2026looktwice}
Marco Morini, Sara Sarto, Marcella Cornia, and Lorenzo Baraldi.
\newblock Look twice: Training-free evidence highlighting in multimodal large language models.
\newblock \emph{arXiv preprint arXiv:2604.01280}, 2026.

\bibitem[Munasinghe et~al.(2025)Munasinghe, Gani, Zhu, Cao, Xing, Khan, and Khan]{munasinghe2024videoglamm}
Shehan Munasinghe, Hanan Gani, Wenqi Zhu, Jiale Cao, Eric Xing, Fahad~Shahbaz Khan, and Salman Khan.
\newblock Videoglamm: A large multimodal model for pixel-level visual grounding in videos.
\newblock In \emph{Proceedings of the IEEE/CVF Conference on Computer Vision and Pattern Recognition}, pages 19036--19046, 2025.

\bibitem[Nie et~al.(2026{\natexlab{a}})Nie, Liu, Zhu, Fan, and Han]{nie2026costeffective}
Frank Nie, Ethan~B. Liu, Yuan Zhu, Wei Fan, and Jindong Han.
\newblock A cost-effective multimodal llm reasoning framework for question answering over irregular clinical time series, 2026{\natexlab{a}}.
\newblock \url{https://arxiv.org/abs/2607.25947}.

\bibitem[Nie et~al.(2026{\natexlab{b}})Nie, Liu, Zhu, Yan, Fan, and Han]{nie2026clirbench}
Frank Nie, Ethan~B. Liu, Yuan Zhu, Loe Yan, Wei Fan, and Jindong Han.
\newblock Clir-bench: Benchmarking multimodal question answering over irregular clinical time series, 2026{\natexlab{b}}.
\newblock \url{https://arxiv.org/abs/2607.09880}.

\bibitem[Ouyang et~al.(2026)Ouyang, Liu, Yao, Cai, Zhou, Meng, Zhou, and Sun]{ouyang2025conan}
Kun Ouyang, Yuanxin Liu, Linli Yao, Yishuo Cai, Hao Zhou, Fandong Meng, Jie Zhou, and Xu~Sun.
\newblock Conan: Progressive learning to reason like a detective over multi-scale visual evidence.
\newblock In \emph{Proceedings of the IEEE/CVF Conference on Computer Vision and Pattern Recognition (CVPR)}, pages 41089--41099, June 2026.

\bibitem[Peng et~al.(2023)Peng, Wang, Dong, Hao, Huang, Ma, and Wei]{peng2023kosmos2}
Zhiliang Peng, Wenhui Wang, Li~Dong, Yaru Hao, Shaohan Huang, Shuming Ma, and Furu Wei.
\newblock Kosmos-2: Grounding multimodal large language models to the world.
\newblock \emph{arXiv preprint arXiv:2306.14824}, 2023.

\bibitem[Rasheed et~al.(2024)Rasheed, Maaz, Mullappilly, Shaker, Khan, Cholakkal, Anwer, Xing, Yang, and Khan]{rasheed2023glamm}
Hanoona Rasheed, Muhammad Maaz, Sahal~Shaji Mullappilly, Abdelrahman Shaker, Salman Khan, Hisham Cholakkal, Rao~M. Anwer, Erix Xing, Ming-Hsuan Yang, and Fahad~S. Khan.
\newblock {GLaMM}: Pixel grounding large multimodal model.
\newblock In \emph{Proceedings of the IEEE/CVF Conference on Computer Vision and Pattern Recognition}, pages 13009--13018, 2024.

\bibitem[Ren et~al.(2025)Ren, Ma, Yang, Wei, Zhang, and Chen]{ren2025vamba}
Weiming Ren, Wentao Ma, Huan Yang, Cong Wei, Ge~Zhang, and Wenhu Chen.
\newblock Vamba: Understanding hour-long videos with hybrid mamba-transformers.
\newblock In \emph{Proceedings of the IEEE/CVF International Conference on Computer Vision}, pages 21197--21208, 2025.

\bibitem[Shen et~al.(2025)Shen, Xiong, Zhao, Wu, Chen, Zhu, Liu, Xiao, Varadarajan, Bordes, et~al.]{shen2024longvu}
Xiaoqian Shen, Yunyang Xiong, Changsheng Zhao, Lemeng Wu, Jun Chen, Chenchen Zhu, Zechun Liu, Fanyi Xiao, Balakrishnan Varadarajan, Florian Bordes, et~al.
\newblock Longvu: Spatiotemporal adaptive compression for long video-language understanding.
\newblock In \emph{Proceedings of the 42nd International Conference on Machine Learning}, 2025.

\bibitem[Sheng et~al.(2025)Sheng, Hao, Li, Wang, and He]{sheng2025sevices}
Yuan Sheng, Yanbin Hao, Chenxu Li, Shuo Wang, and Xiangnan He.
\newblock {SeViCES}: Unifying semantic-visual evidence consensus for long video understanding.
\newblock \emph{arXiv preprint arXiv:2510.20622}, 2025.

\bibitem[Shu et~al.(2025)Shu, Liu, Zhang, Qin, Zhou, Liang, Huang, and Zhao]{shu2025video}
Yan Shu, Zheng Liu, Peitian Zhang, Minghao Qin, Junjie Zhou, Zhengyang Liang, Tiejun Huang, and Bo~Zhao.
\newblock Video-xl: Extra-long vision language model for hour-scale video understanding.
\newblock In \emph{Proceedings of the Computer Vision and Pattern Recognition Conference}, pages 26160--26169, 2025.

\bibitem[Tang et~al.(2026)Tang, Han, Sun, Zhou, Zhang, Wei, Yuan, Xu, and Sun]{tang2026tspo}
Canhui Tang, Zifan Han, Hongbo Sun, Sanping Zhou, Xuchong Zhang, Xin Wei, Ye~Yuan, Jinglin Xu, and Hao Sun.
\newblock {TSPO}: Temporal sampling policy optimization for long-form video language understanding.
\newblock In \emph{Proceedings of the AAAI Conference on Artificial Intelligence}, volume~40, pages 9368--9376, 2026.
\newblock \url{https://arxiv.org/abs/2508.04369}.

\bibitem[Wang et~al.(2024)Wang, Bai, Nah, Wang, Zhang, Chen, Wu, Islam, Liu, and Ren]{wang2024surgicallvlm}
Guankun Wang, Long Bai, Wan~Jun Nah, Jie Wang, Zhaoxi Zhang, Zhen Chen, Jinlin Wu, Mobarakol Islam, Hongbin Liu, and Hongliang Ren.
\newblock Surgical-lvlm: Learning to adapt large vision-language model for grounded visual question answering in robotic surgery, 2024.
\newblock \url{https://arxiv.org/abs/2405.10948}.

\bibitem[Wang et~al.(2026)Wang, Wu, Lu, Liu, Jia, Liu, Nie, Zhu, and Zhang]{wang2026morethinkdeeperqueryexpanded}
Shuning Wang, Zhiheng Wu, YiNuo Lu, Naiming Liu, Chen Jia, Bowen Liu, Shuo Nie, Weijie Zhu, and Yumeng Zhang.
\newblock See more, think deeper: Query-expanded visual evidence and answer-clue guided reflection for long video understanding, 2026.
\newblock \url{https://arxiv.org/abs/2606.09064}.

\bibitem[Wang et~al.(2025)Wang, He, Hong, Cheng, Zhang, Qi, Ding, Gu, Huang, Xu, et~al.]{wang2025lvbench}
Weihan Wang, Zehai He, Wenyi Hong, Yean Cheng, Xiaohan Zhang, Ji~Qi, Ming Ding, Xiaotao Gu, Shiyu Huang, Bin Xu, et~al.
\newblock {LVBench}: An extreme long video understanding benchmark.
\newblock In \emph{Proceedings of the IEEE/CVF International Conference on Computer Vision}, pages 22958--22967, 2025.

\bibitem[Wu et~al.(2024)Wu, Li, Chen, and Li]{wu2024longvideobench}
Haoning Wu, Dongxu Li, Bei Chen, and Junnan Li.
\newblock {LongVideoBench}: A benchmark for long-context interleaved video-language understanding.
\newblock In \emph{Advances in Neural Information Processing Systems}, volume~37, pages 28828--28857, 2024.

\bibitem[Wu et~al.(2025)Wu, Zhang, Wu, Torr, and Gu]{wu2025postalign}
Yixuan Wu, Yang Zhang, Jian Wu, Philip Torr, and Jindong Gu.
\newblock Postalign: Multimodal grounding as a corrective lens for mllms.
\newblock \emph{arXiv preprint arXiv:2506.17901}, 2025.

\bibitem[Xiao et~al.(2026)Xiao, Liu, Liao, Zhang, Lan, Wei, Zhao, Wang, Gu, Ye, Wang, and Xu]{xiao2026vigil}
Xi~Xiao, Chen Liu, Chih-Ting Liao, Yunbei Zhang, Qizhen Lan, Yuxiang Wei, Lin Zhao, Janet Wang, Jianyang Gu, Muchao Ye, Tianyang Wang, and Hao Xu.
\newblock Staying {VIGIL}ant: Mitigating visual laziness via counterfactual visual alignment in mllms.
\newblock \emph{arXiv preprint arXiv:2606.26387}, 2026.

\bibitem[Xie et~al.(2026)Xie, Chen, Ge, and Ni]{xie2026videomtr}
Yuan Xie, Tianshui Chen, Zheng Ge, and Lionel Ni.
\newblock Video-{MTR}: Reinforced multi-turn reasoning for long video understanding.
\newblock In \emph{Forty-third International Conference on Machine Learning}, 2026.
\newblock \url{https://openreview.net/forum?id=UhPwL6LYOc}.

\bibitem[Yan et~al.(2024)Yan, Wang, Yan, Jiang, Hu, Kang, Xie, and Gavves]{yan2024visa}
Cilin Yan, Haochen Wang, Shilin Yan, Xiaolong Jiang, Yao Hu, Guoliang Kang, Weidi Xie, and Efstratios Gavves.
\newblock {VISA}: Reasoning video object segmentation via large language models.
\newblock \emph{arXiv preprint arXiv:2407.11325}, 2024.

\bibitem[Yang et~al.(2026)Yang, Wang, Zhang, Wu, Leng, Zhang, Li, Qin, Lu, Li, et~al.]{yang2026longvt}
Zuhao Yang, Sudong Wang, Kaichen Zhang, Keming Wu, Sicong Leng, Yifan Zhang, Bo~Li, Chengwei Qin, Shijian Lu, Xingxuan Li, et~al.
\newblock Longvt: Incentivizing thinking with long videos via native tool calling.
\newblock In \emph{Proceedings of the IEEE/CVF Conference on Computer Vision and Pattern Recognition}, pages 33816--33826, 2026.

\bibitem[You et~al.(2023)You, Zhang, Gan, Du, Zhang, Wang, Cao, Chang, and Yang]{you2023ferret}
Haoxuan You, Haotian Zhang, Zhe Gan, Xianzhi Du, Bowen Zhang, Zirui Wang, Liangliang Cao, Shih-Fu Chang, and Yinfei Yang.
\newblock Ferret: Refer and ground anything anywhere at any granularity.
\newblock \emph{arXiv preprint arXiv:2310.07704}, 2023.

\bibitem[Zeng et~al.(2025)Zeng, Li, Wang, Li, Jiang, Yan, Li, Shi, Yue, Wang, et~al.]{zeng2025timesuite}
Xiangyu Zeng, Kunchang Li, Chenting Wang, Xinhao Li, Tianxiang Jiang, Ziang Yan, Songze Li, Yansong Shi, Zhengrong Yue, Yi~Wang, et~al.
\newblock Timesuite: Improving mllms for long video understanding via grounded tuning.
\newblock In \emph{International Conference on Learning Representations}, volume 2025, pages 38057--38081, 2025.

\bibitem[Zhang et~al.(2024{\natexlab{a}})Zhang, Zhang, Li, Zeng, Yang, Zhang, Wang, Tan, Li, and Liu]{zhang2024long}
Peiyuan Zhang, Kaichen Zhang, Bo~Li, Guangtao Zeng, Jingkang Yang, Yuanhan Zhang, Ziyue Wang, Haoran Tan, Chunyuan Li, and Ziwei Liu.
\newblock Long context transfer from language to vision.
\newblock \emph{arXiv preprint arXiv:2406.16852}, 2024{\natexlab{a}}.

\bibitem[Zhang et~al.(2026)Zhang, Lin, Wang, Jin, Ding, Ma, and Yang]{zhang2026vticot}
Shufan Zhang, Ziyue Lin, Bairun Wang, Lei Jin, Xuanding Ding, Xinzhu Ma, and Kunlin Yang.
\newblock {VTI-CoT}: Visual-textual interleaved chain of thought for video reasoning.
\newblock \emph{arXiv preprint arXiv:2606.05736}, 2026.

\bibitem[Zhang et~al.(2024{\natexlab{b}})Zhang, Li, Liu, Lee, Gui, Fu, Feng, Liu, and Li]{zhang2024llavanextvideo}
Yuanhan Zhang, Bo~Li, haotian Liu, Yong~jae Lee, Liangke Gui, Di~Fu, Jiashi Feng, Ziwei Liu, and Chunyuan Li.
\newblock Llava-next: A strong zero-shot video understanding model, April 2024{\natexlab{b}}.
\newblock \url{https://llava-vl.github.io/blog/2024-04-30-llava-next-video/}.

\bibitem[Zhou et~al.(2025)Zhou, Shu, Zhao, Wu, Liang, Xiao, Qin, Yang, Xiong, Zhang, et~al.]{zhou2025mlvu}
Junjie Zhou, Yan Shu, Bo~Zhao, Boya Wu, Zhengyang Liang, Shitao Xiao, Minghao Qin, Xi~Yang, Yongping Xiong, Bo~Zhang, et~al.
\newblock {MLVU}: Benchmarking multi-task long video understanding.
\newblock In \emph{Proceedings of the IEEE/CVF Conference on Computer Vision and Pattern Recognition}, pages 13691--13701, 2025.

\bibitem[Zhou et~al.(2026)Zhou, Hua, Jin, Huang, and Duan]{zhou2026reason}
Yuan Zhou, Litao Hua, Shilong Jin, Wentao Huang, and Haoran Duan.
\newblock {ReaSon}: Reinforced causal search with information bottleneck for video understanding.
\newblock In \emph{Proceedings of the AAAI Conference on Artificial Intelligence}, 2026.

\bibitem[Zhu et~al.(2026)Zhu, Chen, Zafeiriou, and Deng]{zhu2026visualflip}
Didi Zhu, Changrui Chen, Stefanos Zafeiriou, and Jiankang Deng.
\newblock {VisualFLIP}: Do predictions depend on task-critical visual evidence in multimodal reasoning?
\newblock \emph{arXiv preprint arXiv:2606.07872}, 2026.

\bibitem[Zou et~al.(2026)Zou, Huang, Zhang, Zhang, and Shen]{zou2026videobrain}
Junbo Zou, Ziheng Huang, Shengjie Zhang, Liwen Zhang, and Weining Shen.
\newblock Videobrain: Learning adaptive frame sampling for long video understanding.
\newblock \emph{arXiv preprint arXiv:2602.04094}, 2026.

\end{thebibliography}

\clearpage
\beginappendix
\section{Extended Related Work}

\noindent\textbf{Long-video understanding.} General video instruction-tuned models such as Video-LLaVA, ShareGPT4Video, LLaVA-NeXT-Video, VideoLLaMA2, VideoChat2, LLaVA-OneVision, and InternVL2 extend image-language alignment and instruction tuning to video inputs~\cite{lin2024video,chen2024sharegpt4video,zhang2024llavanextvideo,cheng2024videollama,li2024mvbench,li2024llava,chen2024expanding}. Long-context models further increase temporal coverage by transferring language-model context length to video or scaling the system around long visual sequences~\cite{zhang2024long,chen2025longvila}. Compact-context methods reduce the cost of long videos through frame- or token-level compression: LLaMA-VID represents each frame with a small number of tokens~\cite{li2024llama}, LongVU performs spatio-temporal adaptive compression~\cite{shen2024longvu}, Video-XL and Video-XL-Pro summarize or reconstruct compressed visual tokens for hour-scale videos~\cite{shu2025video,liu2025video}, and Vamba uses hybrid sequence modeling for hour-long inputs~\cite{ren2025vamba}. Adaptive selection and input-organization methods further improve the input through grounded long-video tuning~\cite{zeng2025timesuite}, query-type-aware frame selection~\cite{li2026divide}, temporal sampling policy optimization~\cite{tang2026tspo}, adaptive frame sampling~\cite{zou2026videobrain}, causal keyframe search~\cite{zhou2026reason}, event-aware frame selection~\cite{chen2026efs}, and panel-style visual repacking~\cite{doorenbos2026videopanels}. Reasoning-oriented methods add native tool calling~\cite{yang2026longvt}, reinforcement-learned video reasoning~\cite{feng2025videor1}, reinforced multi-turn segment selection~\cite{xie2026videomtr,wang2026morethinkdeeperqueryexpanded}, multi-scale visual evidence reasoning~\cite{ouyang2025conan}, and visual-textual interleaved chain-of-thought~\cite{zhang2026vticot}. These works motivate \ours{} but leave a different interface under-specified: how selected frames should be converted into evidence used by the generator.

\noindent\textbf{Evidence grounding and evidence use.} Evidence grounding in MLLMs spans region-level grounding, pixel-level grounding, temporal grounding, and evidence-use diagnostics. Image-based grounded MLLMs such as Kosmos-2, Shikra, and Ferret incorporate bounding boxes, coordinates, or region representations into multimodal dialogue~\cite{peng2023kosmos2,chen2023shikra,you2023ferret}. Pixel-level grounding methods such as GLaMM, GroundingGPT, and LISA generate dense grounded responses or segmentation masks~\cite{rasheed2023glamm,li2024groundinggpt,lai2023lisa}, while PixelEyes casts fine-grained image evidence localization as an active visual search problem~\cite{gong2026pixeleyes}. VISA and VideoGLaMM extend grounding to video objects and pixel-level video regions~\cite{yan2024visa,munasinghe2024videoglamm}. Evidence-centric evaluation and inference methods make evidence dependence explicit: VisualFLIP tests answer changes under task-critical visual edits~\cite{zhu2026visualflip}, VISTAQA and CiteVQA pair answers with pixel-level or element-level evidence~\cite{azadani2026vistaqa,ma2026citevqa}, and Look Twice highlights visual and textual evidence at inference time~\cite{morini2026looktwice}. Diagnostic or corrective work further shows that MLLMs can rely on language or knowledge priors instead of visual evidence~\cite{asadi2026mirage,benlevi2026mirageprobes,jia2026decoding,xiao2026vigil,wu2025postalign}. For video, Moment-Video diagnoses whether models preserve brief answer-critical events~\cite{liu2026moment}, EVIDENT uses entity-grounded visual evidence for cross-domain temporal grounding~\cite{ahn2026evident}, SemVID preserves evidence chains during token pruning~\cite{li2026keeping}, and SeViCES combines semantic and visual evidence for long-video reasoning~\cite{sheng2025sevices}. Closest to latent evidence, RIS grounds image-level latent reasoning tokens with spatial-semantic supervision~\cite{cui2026ris}. \ours{} constructs latent evidence from selected video frames through video-question-answer supervision and uses allocation dispersion for invocation.

Beyond general-domain video, evidence-centered modeling also arises in
biomedical applications. Medical MLLMs established image-centric instruction
tuning and interleaved few-shot reasoning~\cite{li2023llavamed,moor2023medflamingo},
while Surgical-LVLM made answer grounding explicit in surgical scenes~\cite{wang2024surgicallvlm}.
The evidence-use problem becomes sharper over long temporal records: MedHorizon
evaluates sparse evidence in full-length procedures, and Divide-then-Diagnose
organizes ultra-long endoscopy videos into coherent contexts before aggregating
multi-frame findings~\cite{du2026medhorizon,liu2026dividethendiagnose}. In
irregular clinical time series, CLIR-Bench links questions to explicit temporal
evidence, whereas ClinPRISM distills multi-scale observations into compact
LLM-compatible tokens~\cite{nie2026clirbench,nie2026costeffective}. These studies
extend evidence grounding from spatial regions to long, heterogeneous temporal
contexts, reinforcing the need to organize localized observations into
representations that generators can effectively use.

\section{Additional Results}

\subsection{Task Groups for Evidence-Distribution Analysis}
\label{app:rq3_task_groups}

\paragraph{Task taxonomy for RQ3.}
We organize tasks by expected evidence distribution rather than by accuracy. For MLVU, we follow its official taxonomy~\cite{zhou2025mlvu}: holistic LVU uses global information from the whole video and is assigned to distributed evidence, single-detail LVU relies on one critical plot and is assigned to localized evidence, and multi-detail LVU is kept as an intermediate category. For LVBench, we select two representative task families based on human-annotated time-reference spans: summarization questions cover much longer spans than entity-recognition questions and are treated as broad-span evidence, while entity-recognition questions are treated as focal-span evidence.

\subsection{Qualitative Case Studies}
\label{app:case_studies}

\begin{figure}[H]
    \centering
    \includegraphics[width=0.96\textwidth]{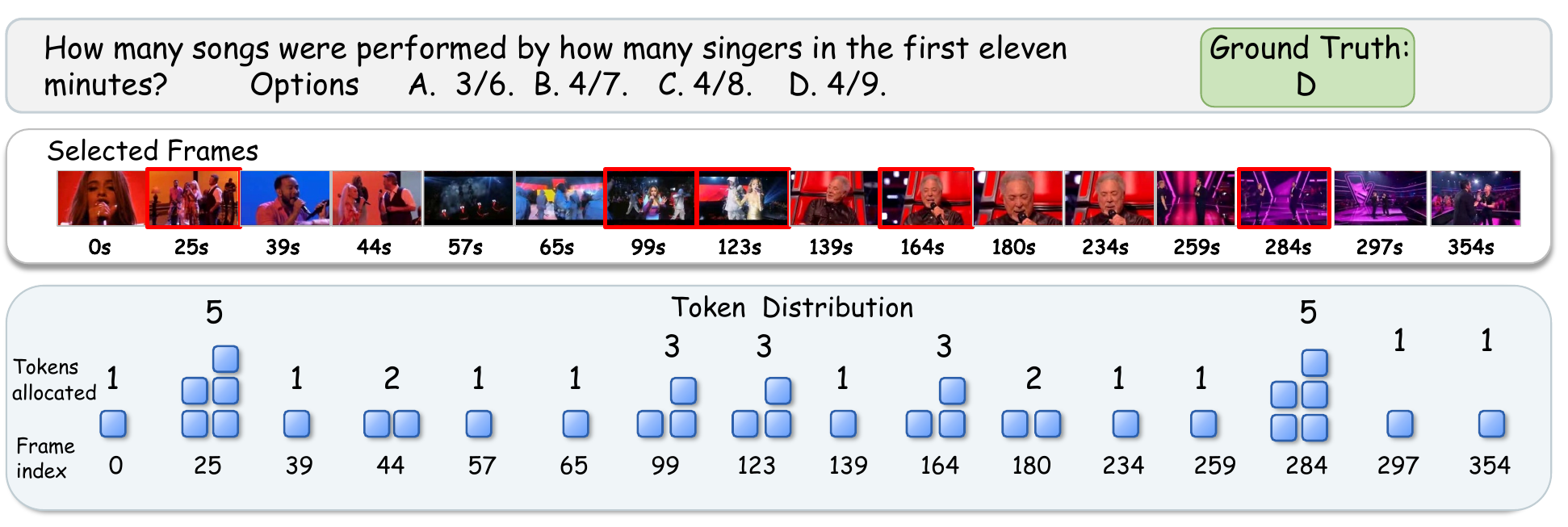}
    \caption{Case 1: Distributed evidence. The latent-token budget is spread
    across multiple supports needed to count the songs and singers.}
    \label{fig:case_distributed}
\end{figure}

\begin{figure}[H]
    \centering
    \includegraphics[width=0.96\textwidth]{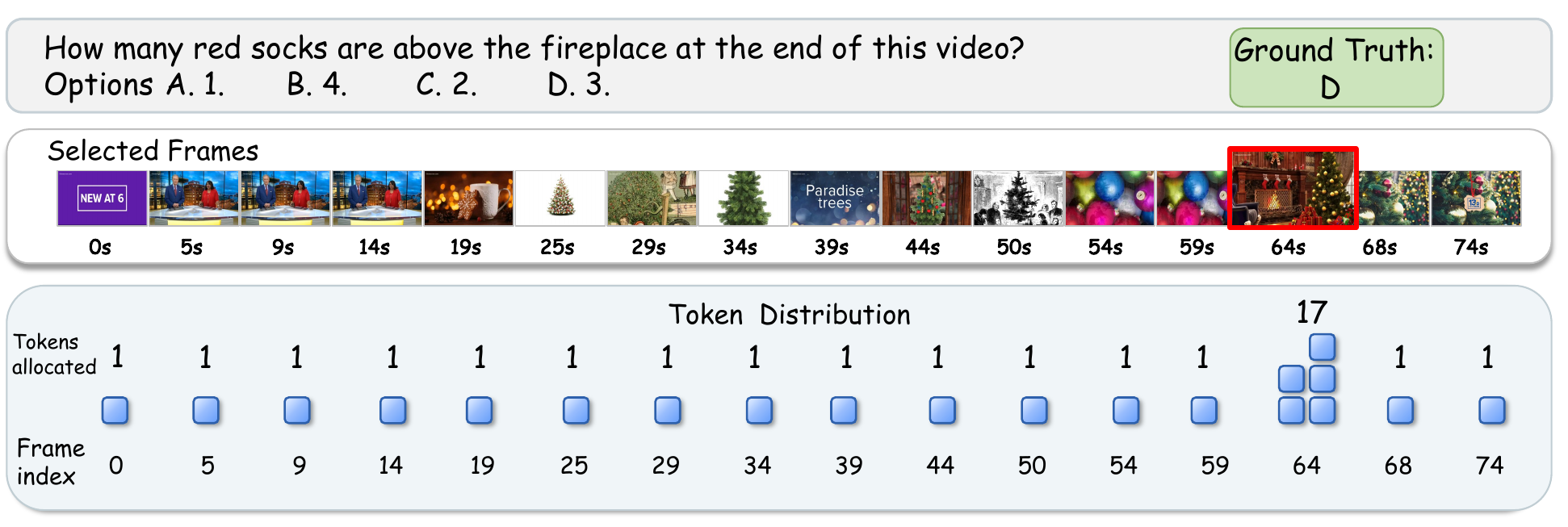}
    \caption{Case 2: Localized evidence. The allocation concentrates on the
    frame containing the answer-critical fireplace.}
    \label{fig:case_localized}
\end{figure}

Figure~\ref{fig:case_distributed} shows a distributed-evidence example. Answering
the counting question requires combining performer information across several
moments, and the allocator assigns additional capacity to multiple corresponding
supports. Figure~\ref{fig:case_localized} shows the complementary localized
pattern. The answer is visible in a single fireplace frame, which receives 17
of the 32 latent tokens, while the remaining frames retain the one-token floor.

\subsection{Token Accounting}
\label{app:token_accounting}

Table~\ref{tab:token_budget} reports visual tokens, average latent evidence
tokens included in the language-model input, and average total video-token cost.
A Latent Evidence Route example contributes exactly $32$ latent tokens, whereas
an Explicit Evidence Route example contributes zero. Shared text prompt tokens
are excluded because they are identical under matched settings. Runtime effects,
including adapter computation, are reported in
Table~\ref{tab:inference_efficiency}. For consistent paper-level accounting, we
count each LLaVA-Video and Qwen2.5-VL frame as $679$ and $768$ visual tokens,
respectively, following their reported backbone conventions.

\begin{table}[H]
\centering
{\small
\setlength{\tabcolsep}{5.5pt}
\renewcommand{\arraystretch}{1.10}
\resizebox{0.5\linewidth}{!}{%
\begin{tabular}{@{}lrrrrr@{}}
\toprule
\textbf{Setting} & \textbf{Frames} & \textbf{Cost/frame} & \textbf{Visual} &
\textbf{Avg. latent} & \textbf{Avg. total} \\
\midrule
\multicolumn{6}{@{}l}{\textit{LLaVA-Video}} \\
Uniform & 8 & 679 & 5,432 & 0 & 5,432 \\
Uniform & 16 & 679 & 10,864 & 0 & 10,864 \\
\ours{} & 8 & 679 & 5,432 & 9.61 & 5,441.61 \\
\ours{} & 16 & 679 & 10,864 & 12.11 & 10,876.11 \\
\addlinespace[2pt]
\multicolumn{6}{@{}l}{\textit{Qwen2.5-VL}} \\
Uniform & 8 & 768 & 6,144 & 0 & 6,144 \\
Uniform & 16 & 768 & 12,288 & 0 & 12,288 \\
\ours{} & 8 & 768 & 6,144 & 24.83 & 6,168.83 \\
\ours{} & 16 & 768 & 12,288 & 24.24 & 12,312.24 \\
\bottomrule
\end{tabular}%
}
}
\caption{Video-token accounting.}
\label{tab:token_budget}
\end{table}

\end{document}